\documentclass[preprint,12pt,authoryear]{elsarticle}

\usepackage{amssymb}
\usepackage{amsmath}

\usepackage{booktabs}  
\usepackage{array}     
\usepackage{makecell}  
\usepackage{color,soul}
\usepackage{enumitem}
\usepackage{comment}
\usepackage{linguex} 
\usepackage{bibentry} 
\usepackage{natbib}
\usepackage{comment}
\usepackage{etoolbox}
\usepackage{multibib}
\usepackage{xcolor}
\usepackage{tipa} 
\usepackage{rotating}
\usepackage{tabularx}
\usepackage{mdframed}
\usepackage{subcaption} 

\definecolor{Acolor}{rgb}{.32,.59,.84}  
\usepackage{bbding}
\usepackage{ragged2e}
\usepackage{adjustbox}
\newcolumntype{P}[1]{>{\RaggedRight\hspace{0pt}}p{#1}}
\usepackage{hanging}
\usepackage{float}
\usepackage{subcaption}
\usepackage[T1]{fontenc}
\usepackage[utf8]{inputenc}
\usepackage{multirow}

\usepackage{xurl}

\journal{Speech Communication}

\begin{document}

\begin{frontmatter}



\title{A Sociolinguistic Analysis of Automatic Speech Recognition Bias in Newcastle English} 

\author[label1]{Dana Serditova}
\affiliation[label1]{organization={University of Regensburg},
            addressline={Faculty of Languages, Literature, and Cultures, Department of English and American Studies},
            city={Regensburg},
            postcode={93040},
            state={Bavaria},
            country={Germany}}

\author[label2,label3]{Kevin Tang\corref{cor1}}
\affiliation[label2]{organization={Heinrich Heine University D\"{u}sseldorf},
            addressline={
            Faculty of Arts and Humanities,
            Institute of English and American Studies,
            Department English Language and Linguistics},
            city={D\"{u}sseldorf},
            postcode={40225},
            state={North Rhine-Westphalia},
            country={Germany}}
\cortext[cor1]{Corresponding author at: Heinrich Heine University D\"{u}sseldorf, Faculty of Arts and Humanities, Institute of English and American Studies, Department English Language and Linguistics, D\"{u}sseldorf, 40225, North Rhine-Westphalia, Germany.\\
E-mail address: kevin.tang@hhu.de (Kevin Tang).}
\affiliation[label3]{organization={University of Florida},
            addressline={Department of Linguistics},
            city={Gainesville},
            postcode={32611-5454},
            state={Florida},
            country={U.S.A.}}



\begin{abstract}
Automatic Speech Recognition (ASR) systems are widely used in everyday communication, education, healthcare, and industry, yet their performance remains uneven across speakers, particularly when dialectal variation diverges from the mainstream accents represented in training data. This study investigates ASR bias through a sociolinguistic analysis of Newcastle English, a regional variety of North-East England that has been shown to challenge current speech recognition technologies. Using spontaneous speech from the Diachronic Electronic Corpus of Tyneside English (DECTE), we evaluate the output of a state-of-the-art commercial ASR system and conduct a fine-grained analysis of more than 3,000 transcription errors. Errors are classified by linguistic domain and examined in relation to social variables including gender, age, and socioeconomic status. In addition, an acoustic case study of selected vowel features demonstrates how gradient phonetic variation contributes directly to misrecognition.

The results show that phonological variation accounts for the majority of errors, with recurrent failures linked to dialect-specific features like vowel quality and glottalisation, as well as local vocabulary and non-standard grammatical forms. Error rates also vary across social groups, with higher error frequencies observed for men and for speakers at the extremes of the age spectrum. These findings indicate that ASR errors are not random but socially patterned and can be explained from a sociolinguistic perspective. Thus, the study demonstrates the importance of incorporating sociolinguistic expertise into the evaluation and development of speech technologies and argues that more equitable ASR systems require explicit attention to dialectal variation and community-based speech data.  
\end{abstract}


\begin{highlights}
\item Systematic ASR errors are strongly linked to dialectal features of Newcastle English.
\item Phonological variation, esp. vowel realisation and glottalisation, drives ASR errors.
\item ASR error rates vary across gender and age within one speech community.
\item Local vocabulary and non-standard grammar trigger ASR errors and standardisation.
\item Speech technologies reproduce existing linguistic and social biases.
\end{highlights}

\begin{keyword}
Automatic Speech Recognition 
\sep 
Newcastle English
\sep gender bias
\sep age bias
\sep socioeconomic bias
\sep dialectal features
\sep error analysis



\end{keyword}

\end{frontmatter}



\section{Introduction}
\label{sec1}

Automatic Speech Recognition (ASR) systems are now routinely used in settings ranging from education \citep{cumbal2024you,butler2019exploration} and healthcare \citep{latif2020speech,adedeji2024sound} to customer service \citep{zou2021design} and everyday personal communication \citep{vacher2010complete, hamalainen2015multilingual}. Despite their growing presence, these systems continue to perform unevenly across speakers, particularly when dealing with dialects that differ from the mainstream norms embedded in their training data \citep{koenecke2020racial, wassink2022uneven, markl2022language, martin2020understanding}. This uneven performance has raised important questions about fairness \citep{liu2022towards}, accessibility \citep{green2021automatic}, and the sociolinguistic assumptions behind the way speech technologies interpret human speech \citep{markl2021context}. When ASR systems consistently struggle to understand a particular group of speakers, it can lead to a negative user experience and the necessity for speakers to accommodate and alter speech patterns \citep{mengesha2021don}. 

A growing body of research demonstrates that these disparities arise not just from technical limitations but also from the linguistic diversity of real-world speech \citep[e.g.,][]{BeraAgarwal2025,dipto-etal-2025-asr,ngueajio2022hey,MartinCorpusData2021}. In the context of automatic speech recognition, \textit{bias} refers to systematic disparities in recognition performance across demographic or linguistic groups. Such disparities are typically quantified by comparing performance metrics across groups, but they do not by themselves identify the source of the disparity, which may instead reflect characteristics of the training data, model architecture, recording conditions, or the evaluation corpus \citep{koenecke2020racial, grieve2025sociolinguistic, lai2023exploring, markl2022language}. Inclusivity gaps were investigated by \citet{ngueajio2022hey}, who developed a benchmark to measure ASR performance across diverse English accents and demonstrated significant disparities, with ASR systems often underperforming for speakers with non-standard accents.  \citet{markl2022language} shows how dialect-specific pronunciation in English can lead to systematic ASR errors, with misrecognitions rooted in phonological differences between regional varieties and Received Pronunciation. Our study takes a wide-scale approach to correlating error patterns with sociolinguistic variation, responding to calls by \citet{markl2022language} for broader sociolinguistic analyses of ASR errors.

This paper examines ASR bias through a focused case study of Newcastle English, a highly recognisable regional variety of North-East England \citep{montgomery2012effect} and one that consistently challenges existing ASR systems \citep{serditova25_interspeech, markl2022language}. While research has shown that dialectal and social factors influence ASR accuracy, there has been little work investigating what specific linguistic features cause ASR errors, or how these errors vary across speakers within the same linguistic community.

Our approach is to treat ASR output as a sociolinguistically patterned behaviour, not just a product of technology, and to show how speech technologies reproduce existing linguistic hierarchies and real-world biases. To do this, the paper combines three strands of analysis: (i) a fine-grained linguistic analysis of more than 3,000 ASR errors, including phonological, morphosyntactic, and lexical errors; (ii) a quantitative assessment of ASR error rates across gender, age, and socioeconomic groups in Newcastle, UK; and (iii) an acoustic case study demonstrating how specific regional vowel realisations directly contribute to misrecognitions. Using naturalistic, spontaneous speech from the Diachronic Electronic Corpus of Tyneside English (DECTE; \citealt{corrigan2012diachronic}) and a state-of-the-art commercial ASR system (Rev AI, \url{https://rev.ai}, selected because it achieved the best overall performance on the DECTE corpus in our previous study \citep{serditova25_interspeech}), we identify recurrent misrecognitions linked to phonological, morphosyntactic, and lexical features of Newcastle English, and test how error rates correlate with social variables such as age, gender, and socioeconomic status. We also incorporate acoustic analysis to illustrate that error patterns can be directly tied to dialect-specific phonetic realisations. Phonological features are initially identified through auditory coding, following established sociophonetic practice. To move beyond categorical labels of linguistic error types and demonstrate that ASR errors are sensitive to gradient phonetic variation, we then incorporate targeted acoustic analysis of selected features. This allows us to show that misrecognitions are not only associated with the presence of dialectal features, but also with fine-grained differences in their phonetic realisation.

These analyses show that ASR systems systematically struggle with salient features of Newcastle English. The resulting errors both arise from socially patterned linguistic variation and, in turn, reinforce the very social and linguistic biases that caused them. Our analysis, grounded in data from the speech community in Newcastle, showcases the value of sociolinguistic expertise in identifying and mitigating bias and offers insights into what building more equitable and dialect-aware ASR systems might look like.

The remainder of the paper is structured as follows: Section~\ref{sec2} reviews existing work on ASR bias; Section~\ref{sec3} outlines the dataset, ASR system selection, and analytical methods; Section~\ref{sec4} presents the quantitative, linguistic, and acoustic results; and Section~\ref{sec5} discusses the implications of these findings for sociolinguistic theory and the development of more equitable ASR systems.

\section{A Range of Biases}
\label{sec2}

In this section, we review existing research on ASR bias across gender and age, ethnoracial affiliation, and dialectal variation. While these categories are relatively well-researched, other sources of bias, such as those based on socioeconomic status (SES), remain largely underexplored and are seldom addressed in ASR evaluation studies \citep{markl2022language, dichristofano2022global}. More frequently, the SES bias is a question of access to technology \citep{mubarak2020confirming, capraro2024impact}, although user experience has been discussed as well \citep{bassignana2025ai}. Yet SES is often linked to speech style and accent strength, with lower SES speakers more likely to use local or stigmatised forms \citep{labov1986social, guy1988language, milroy1993mechanisms, rickford1996regional} that diverge from the linguistic norms embedded in ASR training data. This raises the possibility that SES-related variation in pronunciation or lexical choice could systematically affect ASR performance. In this study, we begin to explore this question directly by examining whether ASR systems exhibit measurable performance differences across socioeconomic groups.

The following subsections outline how some of the more established social and linguistic categories affect ASR performance.

\subsection{Gender and Age Bias}
\label{subsec2.1}

Gender and age bias is a persistent issue in speech recognition. \citet{feng2024towards} studied a range of ASR biases in Dutch and Mandarin. They found that women were generally better recognised than men in Dutch ASR systems, especially for teenagers and older adults. Stronger gender bias was observed in end-to-end (E2E) models compared to hybrid models, even when trained on the same data. The authors attribute this to architectural differences in how models represent speech, noting that ``the way the ASR architectures model the speech also plays a role in inducing bias.'' No significant gender bias was found in Mandarin ASR systems, despite imbalances in training data (however, the training and test data were fully matched in domain). In terms of age, teenagers’ speech was recognised best, while children and older adults were significantly worse recognised, indicating a consistent age-related performance gap in Dutch ASR systems across both architectures and speaking styles. No age-based bias could be evaluated in the Chinese systems due to data constraints.

Previous studies have reported mixed evidence regarding gender-related disparities in ASR performance. For example, \citet{liu2022towards} evaluated four ASR models and found consistently higher WERs for men than for women, with the gender gap reaching up to 45\% relative difference. In contrast, age-related differences were minimal, with no consistent pattern of bias across the 18--85 age range. Fine-tuning the models with in-domain data improved overall recognition accuracy but did not reduce the observed gender disparities.

However, there is work that suggests that these disparities are not consistent across datasets, ASR systems, or speech varieties. Controlled experiments by \citet{garnerin2021investigating} found that overall ASR performance was relatively robust to the gender balance of the training data and appeared to be more strongly influenced by the characteristics of individual speakers included in the training corpus. Similarly, \citet{harris2024modeling} showed that gender-related ASR performance differences depended on the speech variety being recognised: ASR systems produced lower WERs for women’s speech than for men’s speech across several minority English dialects, whereas the opposite pattern was observed for Standard American English.

A large-scale analysis by \citet{jahan2025unveiling} examined twenty ASR systems across four English datasets and found little evidence of systematic gender or age bias. For most systems, gender did not significantly predict error rates, and when bias appeared, it was limited to isolated models and generally disadvantaged men. Age effects were also weak: only a single system showed a statistically significant age-related pattern, and no consistent trend emerged across datasets.

A recent review of ASR fairness has likewise concluded that studies of gender bias report inconsistent findings, with some observing higher error rates for women, others finding no significant gender differences, and still others reporting better recognition for women \citep{patel2025evaluate}.

These studies indicate that evidence for gender- and age-related ASR disparities remains mixed. While some studies report substantial differences in recognition performance across demographic groups, others find little or no significant effects, suggesting that any observed disparities depend on the ASR system, evaluation corpus, and analytical methodology employed.

\subsection{Racial Bias}
\label{subsec2.2}

The topic of racial bias in speech technology has received generous attention from scholars. One of the pioneering papers on the topic was produced by \citet{koenecke2020racial}, who tested the performance of state-of-the-art ASR systems and their ability to transcribe interviews with both White and Black speakers from the US. Ethnoracial affiliation was significant in the systems' performance calculated by Word Error Rate (WER), with Black speakers receiving an almost double error rate compared to their White counterparts. \citet{liu2022towards} reported significant disparities in WERs across different skin tone groups in all ASR systems that were tested within the framework of the study. Similarly, \citet{jahan2025unveiling} found that systems show notable biases related to birthplace, location, native language, and occupation, especially when African American Vernacular English (AAVE) and Mainstream American English (MAE) are compared. 

Possible sources of such a bias are evident when looking into specific linguistic features that deviate from the mainstream from segmental to suprasegmental and from phonetics to morphosyntax. \citet{koenecke2020racial} selected identical phrases spoken by both White and Black speakers. The WER is nonetheless higher for Black speakers. This suggests that linguistic features at the phonetic or phonological level of AAVE could be the cause of the bias. Suprasegmental features have also been shown to play a role in ASR biases. \citet{lai2023exploring} discovered that AAVE speakers experience higher WER particularly when their speech has more variable vowel durations. \citet{MojaradTang_2025_ASRAAE_Interspeech} examined two common phonological features of AAVE, consonant cluster reduction and nasal alveolarisation (ING) and found that their presence in an utterance increase WER. Morphosyntactic features that are absent in Mainstream Englishes were examined by \citet{martin2020understanding} and \citet{heuser24_interspeech}. For instance, \citet{martin2020understanding} focused on habitual ``be,'' a common but unique AAVE feature, concluding that the feature and its surrounding words were more error-prone by ASR than non-habitual ``be''. \citet{johnson2024exploratory} moved beyond binary dialect identification and modelled dialect density in African American English. They demonstrated that both acoustic and grammatical cues contribute to machine perception of dialect.

Thus, previous research makes it clear that ASR misrecognitions often stem from vernacular or dialectal features, which can intersect with ethnoracial affiliation. However, because dialectal variation is not reducible to ethnoracial categories, dialectal bias warrants separate consideration.

\subsection{Dialectal Bias}
\label{subsec2.3}

Existing research has shown that ASR systems struggle with linguistic diversity and underrepresented dialects; here, we concentrate specifically on dialects of English. The \textit{Edinburgh International Accents of English Corpus} (EdAcc) illustrates this clearly, showing significant variation in WER across native English dialects such as Jamaican, Nigerian, Indian, Scottish, and Irish English. While ASR systems like Whisper and a leading commercial model (anonymous) performed reasonably on US and Southern British English, as well as such varieties as South African English and Irish English, they exhibited WERs exceeding 20–30\% on underrepresented varieties like Jamaican and Nigerian English \citep{sanabria2023edinburgh}.

\citet{wassink2022uneven} provide compelling sociophonetic evidence of the issue by evaluating the performance of CLOx -- the University of Washington's sociolinguistic transcription interface that relies on commercial ASR -- across a multi-ethnic speaker sample from the American Pacific Northwest. The study examined Native American (Yakama), African American, ChicanX, and European American speakers, analysing both conversational and read speech. Results showed significantly higher phonetic error rates for non-White ethnic varieties, particularly among Yakama and ChicanX speakers. The paper describes key dialectal features that contributed to ASR errors: (th)-stopping, glottalisation, r-deletion, vowel mergers, and affricate lenition. These features are common in African American and ChicanX English. This study suggests that state-of-the-art ASR systems continue to struggle with dialectal features that are well documented in sociolinguistic research, potentially reflecting their insufficient representation in training data.

\citet{lai2025dialect} tested how well ASR (Dartmouth Linguistic Automation system \citep{reddy2015toward}, an ASR tool based on Hidden Markov Models and Gaussian Mixture Models) handles Appalachian English, a stigmatised and underrepresented dialect of American English. By means of detailed phonetic analysis of ASR transcripts, the paper identified ASR models trained predominantly on mainstream American English give rise to recurrent errors tied to phonological and morphosyntactic features such as vowel mergers, final consonant cluster reduction, and non-standard subject-verb agreement.

A similar endeavour was undertaken in the UK by \citet{markl2022language}, who investigated how algorithmic bias manifests in British English ASR systems. The study demonstrated that commercial ASR systems perform unevenly across regional varieties of British English. In particular, speakers from the North of England and Scotland experienced higher WERs compared to those from the South of England. In this case, the disparities also correlated with well-documented phonological and morphosyntactic features, which may be less common in Standard Southern British English (SSBE) but are salient in regional dialects. The paper argues that these biases are not just technical shortcomings -- instead, they mirror and reproduce broader sociolinguistic hierarchies and language ideologies.

There is a clear underrepresentation in the training data, and several studies have shown that fine-tuning ASR models on a target dialect or training on diversified speech samples indeed leads to improved recognition  \citep{torgbi2025adapting, sanabria2023edinburgh, halpern2022low}. However, as mentioned earlier, performance gaps persist even after fine-tuning \citep{liu2022towards}, which is why it becomes essential to approach the issue from another angle. Rather than only focusing on model optimisation, we argue for a more nuanced linguistic analysis of the specific regional features that cause misrecognitions. By identifying the phonological, lexical, and morphosyntactic patterns that trigger ASR errors, in combination with socioeconomic factors, our work provides a more granular understanding of how linguistic diversity challenges speech technology.

The goal of this study is to highlight how salient local linguistic features can impact ASR performance. To do this effectively, we focus on a specific urban area — Newcastle — where the local dialect (Geordie) is both well-defined and widely recognised by speakers as distinct from other varieties (see Section~\ref{subsec3.3}). Studying a single speech community allows us to examine dialectal variation in context, isolating linguistic features that are systematically underrepresented in ASR training data. We hope to contribute to the conversation about the role of sociolinguistic insight in the development and evaluation of speech technologies. Our research questions are:
\begin{itemize}
    \item[RQ1.]	To what extent do ASR systems misrecognise dialectal features of Newcastle English, and which specific phonological, lexical, and morphosyntactic features are most affected?
    \item[RQ2.]	How do ASR error rates vary across social variables such as gender, age, and socioeconomic status within a single regional speech community?
    \item[RQ3.]	How can sociolinguistic, community-informed analysis identify the potential causes and patterns of bias in speech technologies?
\end{itemize}

This paper builds on \citet{serditova25_interspeech} by substantially expanding both the scope of the dataset and the depth of the analyses. First, we nearly triple the number of manually analysed ASR errors, enabling statistical analyses of the effects of gender, age, and socioeconomic status on ASR performance for the first time. Second, we provide a considerably more detailed qualitative analysis of dialectal errors, documenting the frequency of all error categories, illustrating each with authentic examples, and identifying additional patterns, including the interaction between gender and error type and the effect of socioeconomic status. Third, we extend the taxonomy of ASR errors by introducing analyses of standardisation and spelling errors. Fourth, we complement the qualitative analysis with acoustic analyses of the \textsc{face} and \textsc{goat} vowels to examine the relationship between phonetic realisation and ASR error counts. Finally, we broaden the discussion to consider the role of sociolinguistic research in understanding ASR bias, the relative prominence of different dialectal features, and the interaction of multiple linguistic cues in individual ASR errors.

\section{Method}
\label{sec3}

This section first describes the DECTE dataset and the selection of the ASR system (Sections ~\ref{subsec3.1} and ~\ref{subsec3.2}). We then detail the manual annotation and statistical analysis of ASR errors (Sections ~\ref{subsec3.3} and ~\ref{subsec3.4}), followed by the methods used for the acoustic analysis of \textsc{face} and \textsc{goat} vowels (Section ~\ref{subsec3.5}).

\subsection{Dataset: DECTE Corpus}
\label{subsec3.1}

\textit{Diachronic Electronic Corpus of Tyneside English} (DECTE)  \citep{corrigan2012diachronic} -- a representative corpus of dialect speech from the Tyneside area of North-East England -- serves as the main source of data. DECTE is a sociolinguistic sample representative of the local population and dialectal landscape. The corpus comprises 72 hours of naturalistic, spontaneous speech from 160 speakers (excluding the interviewers and distributed across 99 files) of different ages, genders, socioeconomic backgrounds, and levels of education. All of the speakers come from the Newcastle area. Furthermore, the corpus is fully transcribed by human annotators, including dialectal features. Thus, DECTE provides verified transcriptions of what speakers actually said (the ground truth), allowing us to directly compare ASR output against ground truth.

From this corpus, we selected a subset of 47 recordings containing 81 speakers for analysis. The total duration of the subset was 34 h 00 min 44 s (mean duration = 43 min 25 s, SD = 15 min 00 s, range = 18 min 32 s–1 h 29 min 27 s). Recordings were chosen to obtain a balanced distribution of speakers across age and gender and, where possible, socioeconomic status, while also prioritising recordings with high acoustic quality as measured by the Waveform Amplitude Distribution Analysis Signal-to-Noise Ratio (WADA-SNR, \citet{kim2008robust}). The distribution of participants included in the analyses is provided in Appendix~\ref{tab:participants}. Sampling continued until the dataset yielded approximately 3,000 manually analysed ASR errors, which was considered sufficient to support both the quantitative and qualitative analyses presented in this study.

\subsection{Tool: ASR System Selection}
\label{subsec3.2}

In line with the methodology established in our previous work \citep{serditova25_interspeech}, we applied a two-stage ASR selection process based on clear evaluative criteria. First, we considered only systems that reflect state-of-the-art ASR technology, particularly those leveraging recent advances in deep learning and large-scale speech modelling. Second, we prioritised systems that are commercially available or otherwise accessible to non-specialist users to ensure that our findings remain relevant to real-world applications.

In \citet{serditova25_interspeech}, we pre-tested the DECTE corpus on four ASR systems (Google Cloud Speech-to-Text (\url{https://cloud.google.com/speech-to-text}), CrisperWhisper (an advanced variant of OpenAI's Whisper) \citep{zusag24_interspeech}, Deepgram Voice AI (\url{https://deepgram.com/}), and Rev AI (\url{https://rev.ai})) using a representative 10\% sample. Based on initial performance, two systems (Google and Deepgram) were excluded due to high WERs, while Rev AI and CrisperWhisper were further compared on a larger subset. Rev AI, configured for UK English, consistently outperformed CrisperWhisper across speaker groups, achieving an average WER of 31.95\% on the full dataset. This framework informed the selection of the most robust and dialect-sensitive ASR system for further analysis. We excluded the systems with high WERs because such high error rates would produce a floor effect, making it difficult to isolate which linguistic features cause errors because everything is transcribed inaccurately. Our goal in this selection process was to identify the system that performed best on this dialectal dataset, as our intention was to demonstrate that even the highest-performing system still produces substantial dialect-related errors.

The main downside of working with Rev AI is the fact that it functions as a black box: users have no access to the model architecture, training data, or decision-making processes. And because the underlying model and training data are proprietary, we can not inspect or fine‑tune the system. However, a system like Rev AI is representative of real-world ASR use. A typical ASR user — be it in education, business, or accessibility contexts — is unlikely to have the technical expertise or resources to fine-tune models, nor access to a large, annotated training dataset. By evaluating an off-the-shelf, commercially available system, our goal is to reflect the experience of everyday users who rely on these technologies out of the box, without the ability to adapt them to specific linguistic contexts.

47 recordings containing 81 speakers distributed equally by gender and age and, where possible, socioeconomic background, were processed using Rev AI. WER was then calculated for each file. WER is defined as the number of word substitutions, deletions, and insertions in ASR output, divided by the total number of words in the ground truth (the actual utterance as spoken by the human), and multiplied by 100 \citep{klakow2002testing, hunt1990figures}. In the next step, manual error analysis took place.

\subsection{Error Analysis}
\label{subsec3.3}

The error analysis focused on dialectally meaningful errors associated with established phonological, morphosyntactic, and lexical features of Newcastle English. Tables~\ref{tab:tyneside_features_phon} and~\ref{tab:tyneside_features_syntax} summarise these features together with their Standard Southern British English (SSBE) counterparts \citep{roach2009english, carr2019english, lindsey2019english, gut2009introduction}, which frequently occurred as ASR substitutions in our data. These features formed the basis of the annotation scheme and are used to organise the error analyses presented in Section~\ref{sec4}.

\begin{table}[H]
    \centering
    \caption{Summary of salient Newcastle English features: phonetic and phonological features, with Standard Southern British English counterparts for comparison. WC = working class; MC = middle class.}
    \small
    \begin{tabular}{p{3.3cm}p{5cm}p{4.3cm}}
        \toprule
        \textbf{Feature Name} & \textbf{Newcastle English} & \textbf{Standard Southern British English} \\
        \midrule
        Retention of \textipa{[h]} & 
        Typically retained initially, but dropped among the WC \citep{hughes2013english}. 
        & Typically retained in all phonological environments. \\

        Lack of dark [\textltilde] &  Typically a clear \textipa{[l]} \citep{hughes2013english}. & Syllable-final dark [\textltilde]. \\

        HappY-tensing & Realised as \textipa{[i]} or \textipa{[i:]} \citep{hughes2013english}. & More centralised, closer to \textipa{[I]}. \\

        \textsc{foot/strut} Split Absence & \textsc{foot} and \textsc{strut} vowels both realised as \textipa{[U]} \citep{hughes2013english}. & Distinct vowels: \textsc{foot} \textipa{[U]} or [\textbaro], \textsc{strut} \textipa{[2]}. \\

        Vowel Quality in \textsc{bath} and \textsc{trap} & No \textsc{bath-trap} split, no \textsc{bath} retraction, \textsc{trap} pronounced with [a] \citep{beal2004english, hickey2015north}. 
        & \textsc{bath} words have retracted \textipa{[A:]} (e.g., “path” \textipa{[pA:\texttheta]}). \textsc{trap} pronounced with [\ae]. \\

        Vowel Quality in \textsc{fleece} and \textsc{goose} & Closer to cardinal vowels in closed syllables \citep{mearns2015tyneside}. & Fronted \textsc{goose} [\textbaru] or \textipa{[u:]}; \textsc{fleece} \textipa{[i:]}, possibly diphthongal [\textbaru w], \textipa{[Ij]}. \\

        Vowel Quality in \textsc{face} and \textsc{goat} & Monophthongal realisations; centering diphthongs \textipa{[i@]} and \textipa{[u@]} (older WC); \textsc{goat} [\textbaro] (younger MC) \citep{watt2002don}. & \textsc{face} \textipa{[eI]} and \textsc{goat} \textipa{[@U]} are common diphthongs. \\

        Vowel Quality in \textsc{near} and \textsc{square} &  \textipa{[I\textschwa]} and \textipa{[I\textturna}], as well as \textipa{[\textepsilon\textschwa]} \citep{hughes2013english}. & \textsc{near} \textipa{[I\textschwa]} and \textsc{square} \textipa{[e\textschwa]} are common realisations. \\

        Vowel Quality in \textsc{price} & Realised as \textipa{[EI]}, \textipa{[i:]}, or \textipa{[A:]} \citep{beal2004english}. & Typically realised as \textipa{[aI]}. \\

        Near glottalisation of /p/, /t/, /k/ & Occurs between sonorants \citep{docherty1999sociophonetic}. & Uncommon, except syllable-final pre-consonantal /t/. \\

        ‘g’-dropping (\textit{-ing} \textipa{[In]}) & Common in informal or WC speech; varies with age and context \citep{grama2023post}. & Present in informal speech but avoided in formal registers. \\

        \textsc{nurse} and \textsc{north} Merger & Merged in older WC; \textsc{north} rounded \textipa{[\o]} in younger women \citep{watt2014patterns}. & \textsc{nurse} and \textsc{north} are distinct: \textipa{[3:]} vs.\ \textipa{[O:]}. \\

         T-to-R Rule & E.g., in ``get off'' [g\textepsilon\textturnr of], less common in younger speakers, mostly found in older women \citep{carr1999sociophonetic, beal2004english}. & Mainstream pronunciation with [t] instead of [\textturnr]. \\
       
        \bottomrule
    \end{tabular}
    \label{tab:tyneside_features_phon}
\end{table}

\begin{table}[H]
    \centering
    \caption{Summary of salient Newcastle English features: morphosyntactic and lexical features, with Standard Southern British English counterparts for comparison.}
    \small
    \begin{tabular}{p{3.3cm}p{5cm}p{4.3cm}}
        \toprule
        \textbf{Feature Name} & \textbf{Newcastle English} & \textbf{Standard Southern British English} \\
        \midrule
        
        Unmarked Plurals & “six month”, “three pound” \citep{beal2004english}. & Marked with regular plural -s. \\
        Multiple Negation & Present in Newcastle English \citep{beal1993grammar}. & Considered non-standard. \\
        Conflation of Past and Past Participle & “they’ve broke it” \citep{beal1993grammar}. & Distinction maintained (“they’ve broken it”). \\
        Pronouns & Regional forms ``yous'' (2nd pl.), ``wor'' (meaning ``our"), ``us'' in object position instead of ``me''  \citep{beal1993grammar, beal2004english, pearce2012folk}. & Standard forms ``you'', ``our'' \textipa{[aU@]} or \textipa{[A:]}, ``me''. \\
        Local Vocabulary & \textit{bairn}, \textit{clamming}, etc. \citep{hughes2013english}. & These words are not used. \\
        \bottomrule
    \end{tabular}
    \label{tab:tyneside_features_syntax}
\end{table}

The annotation protocol was adapted from our previous study on the DECTE corpus \citep{serditova25_interspeech}. As in that work, only dialectally meaningful errors — that is, errors attributable to regional phonological, lexical, or morphosyntactic variation — were manually coded, whereas errors resulting from background noise, overlapping speech, or other technical factors were excluded. Errors were identified by aligning ASR output with human-verified transcriptions at the word level using a Python-based string-alignment script\footnote{Word-level alignment between ASR output and the human-verified transcription was performed using the \texttt{difflib} module for sequence comparison, and WER was computed using the \texttt{jiwer} Python library \citep{jitsi_jiwer_2020}.}. Two trained sociophoneticians (the first author and an assistant) manually reviewed the transcripts while listening to the audio. A detailed coding protocol was developed prior to annotation, specifying how each error type and sub-type should be identified and categorised. Errors were categorised at two levels: by error type (e.g., phonological, morphosyntactic, standardisation) and by finer-grained sub-levels (e.g., vowel quality, verb paradigm). At the outset, both annotators independently coded a subset of the data to ensure consistency in applying the scheme; discrepancies were discussed and resolved jointly, and the coding protocol was refined accordingly. After this calibration stage, the remaining data were divided between the annotators. Throughout the process, any unclear or controversial cases were flagged and discussed until consensus was reached. Errors that could not be confidently linked to a dialectal feature or assigned to a specific linguistic category were excluded from the analysis. 

In this study, we expanded our error analysis sample from 32 to 81 speakers distributed across 47 recordings, resulting in 3,005 errors that were classified -- triple the number of errors compared to our previous study. Our analysed dataset contains the following information, allowing us to control for sociolinguistic variables, individual differences in the quality of the audio files, and the linguistic features that were most prone to errors:

\begin{enumerate}
    \item Error type (the linguistic domain): phonological, morphosyntactic, lexical, standardisation, spelling, or plural elision error. Phonological errors were those in which the ASR misrecognition could be attributed to pronunciation (e.g., vowel quality or glottalisation); morphosyntactic errors involved non-standard grammatical forms (e.g., verb paradigms or pronouns); lexical errors involved local vocabulary items (e.g., ``bairn''); standardisation errors involved replacing dialectal features with SSBE forms (e.g., ``me life'' → ``my life'', ``telly'' → ``television''); spelling errors involved incorrect orthographic rendering without a change in lexical identity. Plural elision involved the omission of the final \textit{-s}, e.g., ``dogs'' → ``dog''.
    \item Error specification within the linguistic domain. These refer to broad groups of the local linguistic features, such as vowel quality, `g'-dropping [\textipa{In}], monophthongisation, verb paradigm, conflation of past tenses, etc.
    \item  A specific linguistic feature that the error likely stems from (see Tables~\ref{tab:tyneside_features_phon} and~\ref{tab:tyneside_features_syntax} for an overview).
    
    \item The actual error. Examples of errors will be provided and are given in the following format throughout this paper: ``X''→``Z'', where ``X'' is the ground truth, ``→'' is ``transcribed as'', and ``Z'' is ASR output.    
    \item  Information about the gender, age, education, and occupation of the speaker.    
    \item  Signal to Noise ratio (SNR) of each audio file estimated with the Waveform Amplitude Distribution Analysis (WADA-SNR) \citep{kim2008robust}.
\end{enumerate}

\subsection{Error Analyses: Statistical Modelling of Social Factors}
\label{subsec3.4}

To examine how linguistic error counts varied across demographic variables and audio quality, we fitted a series of generalised linear mixed-effects models (GLMMs) using the \texttt{lme4} package (Version 1.1-38; \citet{lme2015}) in R \citep[Version 4.5.2;][]{r}. The dependent variable was error count. Fixed effects included error type, age group, gender, socioeconomic status, and audio quality. Age group consisted of four categories (16--20, 21--40, 41--60, and 61--90 years), gender comprised men and women, socioeconomic status comprised working-class, middle-class, student, retired, unemployed speakers, and homemakers, and audio quality was represented by the continuous WADA-SNR measure. To account for repeated observations, a random intercept for speaker (participant ID) was included in all models.

Given the count nature of the outcome variable, we initially fitted Poisson GLMMs. However, tests for overdispersion revealed substantial overdispersion (overdispersion ratio = 3.26, $p < .001$), violating Poisson assumptions. As a result, all subsequent analyses were conducted using Negative Binomial GLMMs fitted with \texttt{glmer.nb()}. 

We evaluated alternative fixed-effects structures using likelihood ratio tests via the \texttt{anova()} function, including models with an interaction between error type and gender, error type and age group, and error type and SES. A more complex model including a three-way interaction between error type, gender, and age group failed to converge reliably and was therefore excluded from further consideration. Regarding the random-effects structure, we initially attempted a maximal specification with random slopes for error type by participant $(1 + \textit{error\_type} \mid \textit{participant})$. This model proved statistically unidentifiable because the number of random-effects parameters exceeded the number of observations. Consequently, we retained a simpler random-intercept-only structure.

Age group, socioeconomic status, and audio quality (WADA-SNR) were initially included as fixed effects but were excluded from the final model because they did not significantly improve model fit according to likelihood ratio tests and AIC comparisons. The final best model has the dependent variable \texttt{error\_count}, predicted by the interaction between \texttt{error\_type} and \texttt{gender}, as well as their main effects. A random intercept was included for each participant (\texttt{participant\_id}). Gender and error type were sum-coded using \texttt{contr.sum}, with reference levels set to ``men'' and ``morphosyntax''. The model formula was:

\begin{align*}
\texttt{error\_count} \sim\ & \texttt{error\_type} \times \texttt{gender} \\
& + \texttt{error\_type} + \texttt{gender} \\
& + (1 \mid \texttt{participant\_id})
\end{align*}

Post-hoc model diagnostics were performed to validate the final negative binomial mixed-effects model. A dispersion test indicated no substantial overdispersion (dispersion test: $p$ = .504), and the outlier test revealed no influential points ($p$ = 1). The Kolmogorov–Smirnov test on quantile-scaled residuals also showed no significant deviation from uniformity ($p$ = .433), confirming appropriate model fit. Residual distribution by predicted bins showed homogeneity of variance, as assessed by a Levene test (not significant). Generalised variance inflation factors (GVIFs; \citet{fox1992generalized}) were all well below 2, indicating no multicollinearity issues. Estimated marginal means were computed using the \texttt{emmeans} package \citep{emmeans} to facilitate post-hoc comparisons among factor levels.

\subsection{Acoustic Analyses: \textsc{face}- and \textsc{goat}-monophthongisation}
\label{subsec3.5}

To strengthen the argument that ASR errors are indeed connected to the local linguistic features, we conduct a case study of \textsc{face}- and \textsc{goat}-monophthongisation. The selection of these vowels is motivated by the error analysis: monophthongisation was one of the three leading phonological error subcategories, accounting for 22.1\% of all phonological errors (Table~\ref{tab:phon_errors}). \textsc{face} and \textsc{goat} are particularly suitable for a more fine-grained acoustic investigation because Newcastle English is characterised by salient regional realisations of both vowels, including monophthongal and centering variants (Table~\ref{tab:tyneside_features_phon}). The purpose of this analysis is therefore not to provide a speaker-level decomposition of overall WER, but to move one level below the categorical error analysis and examine acoustically whether gradient phonetic realisation of a recurrent error-producing feature is associated with ASR misrecognition. Specifically, we examine whether speakers with more strongly monophthongal or centering realisations of \textsc{face} and \textsc{goat} also receive more ASR errors associated with these vowels.

To conduct the analysis, we used a combination of time-aligned transcriptions. A subset of the DECTE recordings was manually time-aligned using ELAN \citep{elan2025}. To increase efficiency and extend data coverage, we also incorporated manually aligned DECTE transcription files from the \emph{Language Change Across the Lifespan} project \citep{grama2023post, bauernfeind2023change}, provided by Isabelle Buchstaller and James Grama. These supplemental files, already aligned according to the project's standards, allowed us to reduce the time and resources required for manual alignment while maintaining consistency in transcription quality. For the analysis, we selected 12 speakers representing low, mid-range, and high counts of \textsc{face} and \textsc{goat}-related ASR errors (exact numbers of errors are in Section~\ref{subsec:acousticanalyses}). These speakers were distributed across 7 recordings, 2 of which were transcribed and time-aligned by the first author and an assistant and 5 by the \emph{Language Change Across the Lifespan} project team. The selected speakers contributed between 2,178 and 8,929 words each (mean = 5,867, SD = 2,077). The manually time-aligned ELAN \texttt{.eaf} annotation files were subsequently processed with the Montreal Forced Aligner (MFA; \citet{mcauliffe17_interspeech}) to obtain phone-level segmentations required for automated formant extraction. Vowel formants were then extracted using \texttt{new-fave} \citep{fave_2022}, from which we used the Discrete Cosine Transform (DCT) coefficients in the present analyses.

Previous acoustic work has used F1 and F2 trajectories to quantify diphthongal movement and the degree of monophthongisation \citep{fox2009cross, olsen2017impact, serditova2026sounds}. In the present study, we represent these time-varying formant trajectories using Discrete Cosine Transform (DCT) coefficients, analysed in R \citep{r} following \citet{fruehwald2024}. DCT approximates a formant trajectory using weighted cosine functions, providing a compact representation of its dynamic properties. In particular, the resulting coefficients capture aspects of formant movement such as the magnitude and shape of the glide, making this approach well suited to the analysis of monophthongisation \citep{oppermann2023s, cox2024australian}.

For this case study, we analysed 1,030 \textsc{face} tokens and 827 \textsc{goat} tokens of 12 speakers. We examine and visualise the realisation of these vowels to investigate the potential acoustic and sociolinguistic factors associated with ASR errors.

\section{Results}
\label{sec4}

This section presents the error analyses (Section~\ref{subsec:errordistribution}) and the acoustic analyses (Section~\ref{subsec:acousticanalyses}).

Section~\ref{subsec:errordistribution} reports the error analyses in three stages: (i) descriptive statistics by social factors (Section~\ref{subsubsec4.1.1}), (ii) mixed-effects regression modelling (Section~\ref{subsubsec4.1.2}), and (iii) a detailed descriptive linguistic analysis of phonological, morphosyntactic, lexical, standardisation, and spelling errors (Sections~\ref{subsubsec4.1.3}–\ref{subsubsec4.1.6}).

Section~\ref{subsec:acousticanalyses} presents the acoustic analyses of \textsc{face}- and \textsc{goat}-monophthon-gisation, examining the relationship between the number of diphthong-related ASR errors in a speaker’s speech and the phonetic realisation of these diphthongs.

\subsection{Error Distribution}
\label{subsec:errordistribution}

The distribution of error types observed across the dataset is summarised in Table~\ref{tab:errordistribution}. Phonological errors constitute the most frequent category by a substantial margin, with a total of 1,826 instances (60.8\%). This high frequency suggests that pronunciation-related features were particularly salient and variable in the data. Lexical errors follow with 590 instances (19.6\%), indicating a notable number of issues related to word choice or vocabulary. Standardisation errors (294 instances, 9.8\%) and morphosyntactic errors (235 instances, 7.8\%) occur at comparable rates, indicating that ASR struggles to handle divergence from conventional forms at both the orthographic and grammatical levels. In contrast, spelling errors are relatively rare, with only 21 instances (0.7\%).\footnote{Plural elision errors (n = 39, 1.3\%), where ASR omitted word-final \textit{-s} (e.g., ``flats'' → ``flat'', ``cities'' → ``city''), are excluded from the table due to their unclear linguistic pattern. Since no consistent dialectal explanation emerged, a possible technical cause is that background noise or high-pass filtering may have led to the suppression of high-frequency sounds such as final /s/.}

\begin{table}[H]
\centering
\caption{Frequency distribution of error types.}
\label{tab:errordistribution}
\begin{tabular}{lrr}
\toprule
\textbf{Error Type} & \textbf{Count} & \textbf{Percentage} \\
\midrule
Phonological     & 1,826 & 60.8\% \\
Lexical          & 590  & 19.6\% \\
Standardisation  & 294  & 9.8\%  \\
Morphosyntax     & 235  & 7.8\%  \\
Spelling         & 21   & 0.7\%  \\
\bottomrule
\end{tabular}
\end{table}

\subsubsection{Social Factors: Descriptive Statistics}
\label{subsubsec4.1.1}

We begin the breakdown by reporting errors based on social factors. The ASR system produced more transcription errors for men than for women, with speech from men accounting for 57.4\% of all errors and speech from women accounting for 42.6\%. As for age, younger and older adults received more errors (26.3\% for those between 16 and 20 and 31.4\% for those between 61 and 90) than working-age adults (i.e., speakers aged 21-60; 18.0\% and 24.4\% for the 21-40 and 41-60 age groups, respectively).
This confirms our previous findings that both younger and older speakers experience higher ASR error rates, 
suggesting that the systems perform most accurately for working-age adults \citep{serditova25_interspeech}.

\begin{figure}[H]
\begin{flushleft}
    \centering
    \includegraphics[width=0.8\textwidth]{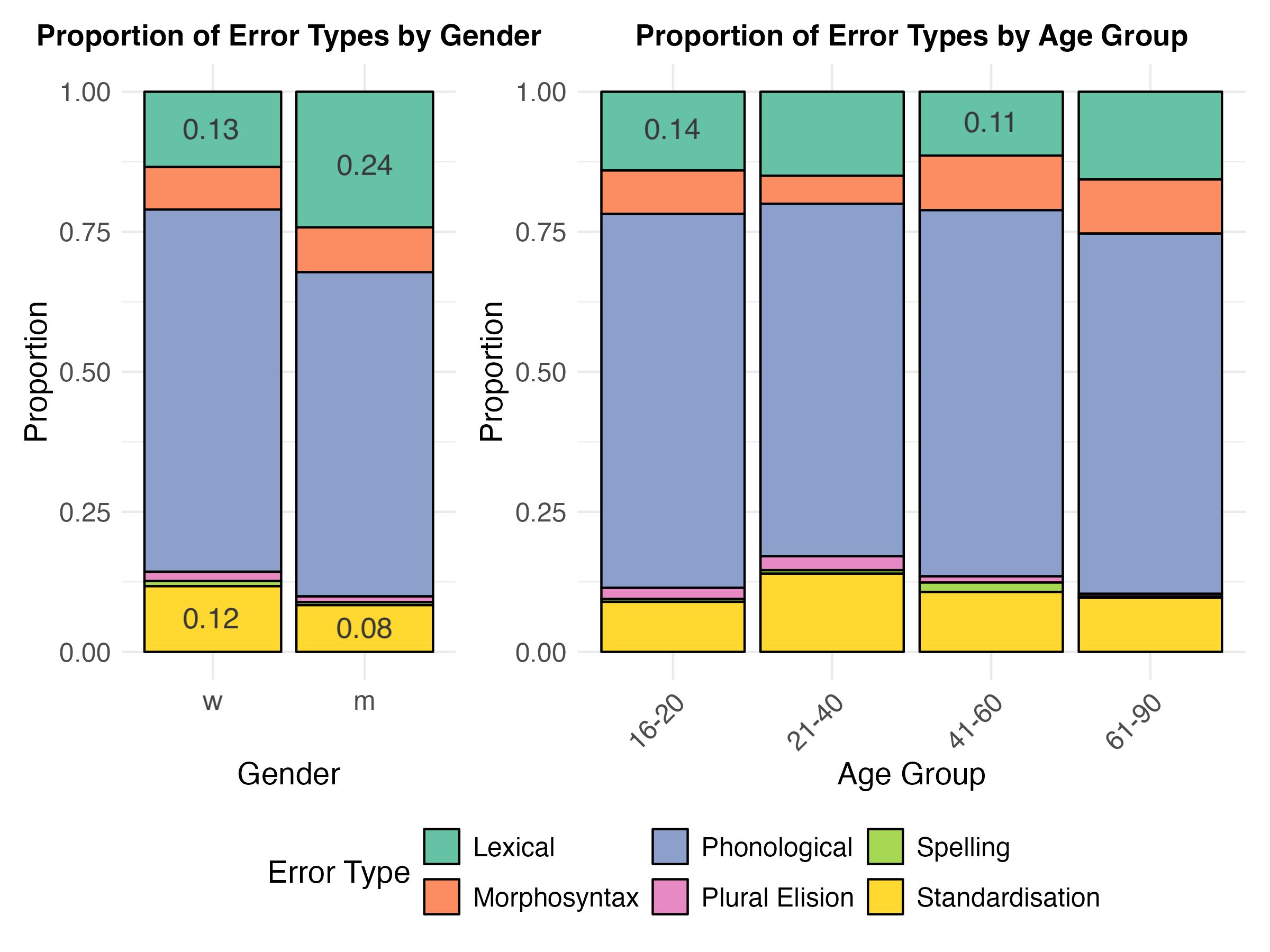}
    \caption{Proportion of error types by gender and age.}
    \label{fig:combined_plot}
\end{flushleft}
\end{figure}

Figure \ref{fig:combined_plot} shows proportions of error types by gender and age. Lexical errors were notably more prominent among men, who received 24\% of lexical errors, compared to 13\% for women. Younger and older speakers received more lexical errors than working-age adults (14\% for those aged 16-20 vs. 11\% for those aged 41-60). Standardisation errors were slightly more persistent for the women in our dataset (12\% for women and 8\% for men). No notable differences can be observed with the other error types.

\begin{figure}[H]
\begin{flushleft}
    \centering
    \includegraphics[width=0.8\textwidth]{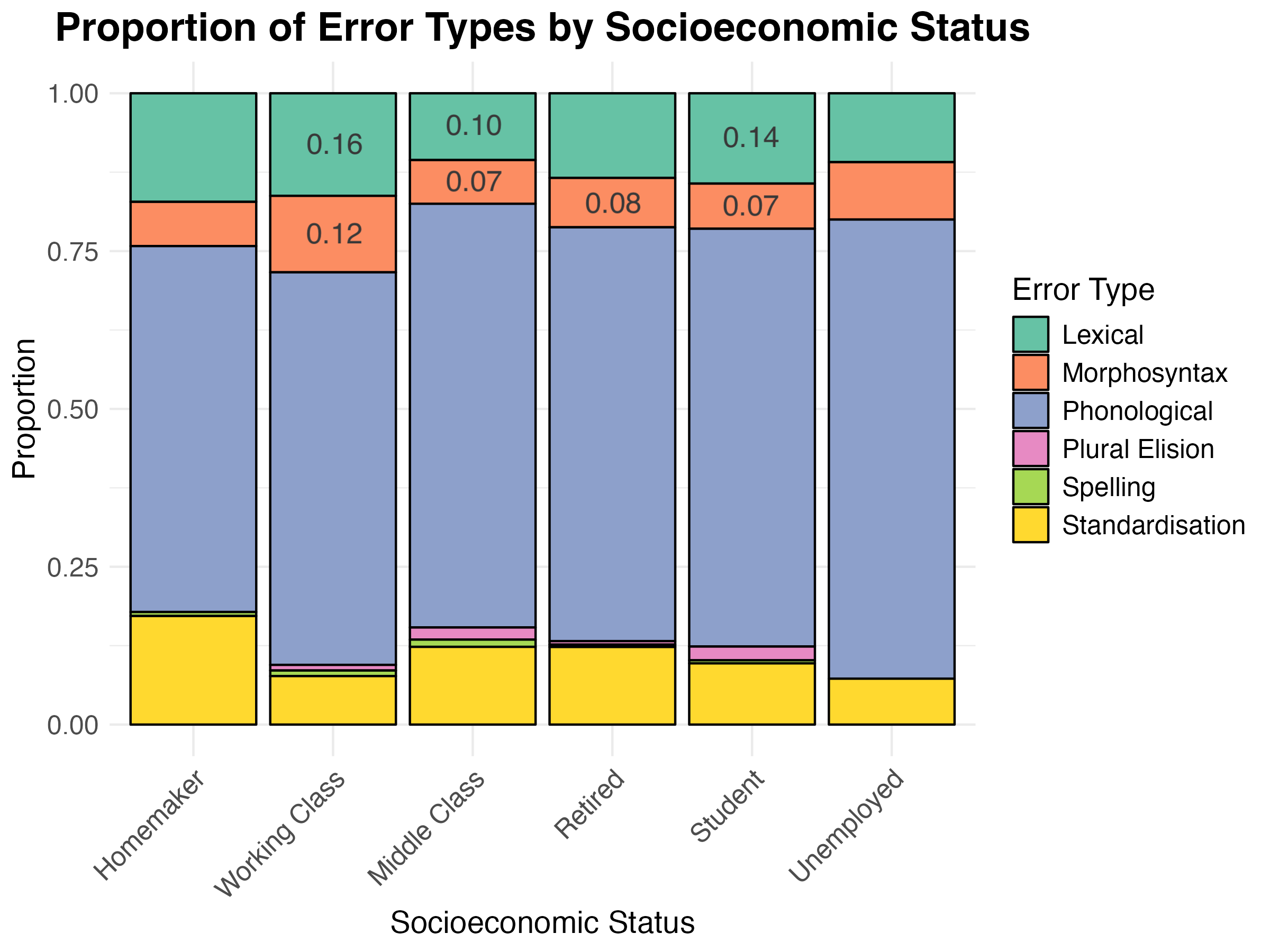}
    \caption{Proportion of error types by socioeconomic status.}
    \label{fig:class_plot}
\end{flushleft}
\end{figure}
  
As for SES, overall, students received the poorest ASR performance (28.4\% of errors), followed by working class (WC) (25.5\%). Middle-class (MC) and retired speakers received 19.3\% and 18.4\% of errors, respectively. Figure \ref{fig:class_plot} demonstrates the distribution of errors by SES. While phonological errors dominate, more differences can again be seen in the lexical domain. WC speakers and students received more lexical errors (16\% and 14\%) than MC speakers (10.5\%). Furthermore, WC speakers received the most morphosyntactic errors (12\%, as opposed to 7\% for both MC and students and 8\% for retired speakers). 
This suggests that ASR systems may align more closely with speech patterns of the MC 
and perform less accurately for WC or student speech. Possible sociolinguistic explanations for these descriptive patterns are discussed in Section~\ref{sec5}.

\subsubsection{Social Factors: Inferential Statistics}
\label{subsubsec4.1.2}

A GLMM with a negative binomial distribution was fitted to predict error counts. The fixed effects included error type and gender, and their interaction, with a random intercept for speaker (Table \ref{tab:stats}).

\begin{table}[H]
\centering
\caption{Fixed effects from the generalised linear mixed model (Negative Binomial) predicting error count. The model includes a random intercept for speaker.}
\begin{tabular}{l r r r r c}
\toprule
\multicolumn{1}{l}{\textbf{Predictor}} &
\multicolumn{1}{c}{\textbf{$\hat{\beta}$}} &
\multicolumn{1}{c}{\textbf{SE}} &
\multicolumn{1}{c}{\textbf{z}} &
\multicolumn{1}{c}{\textbf{p}} &
\multicolumn{1}{c}{\textbf{Sig.}} \\
\midrule
(Intercept) & 1.955 & 0.078 & 25.017 & $< .001$ & *** \\
\texttt{Lexical} & -1.324 & 0.170 & -7.806 & $< .001$ & *** \\
\texttt{Standardisation} & -0.040 & 0.146 & -0.273 & .785 & \\
\texttt{Phonological} & 2.430 & 0.131 & 18.513 & $< .001$ & *** \\
\texttt{Gender: women} & 0.201 & 0.153 & 1.316 & .188 & \\
\texttt{Lexical $\times$ Women} & 0.017 & 0.338 & 0.050 & .960 & \\
\texttt{Standardisation $\times$ Women} & 0.701 & 0.293 & 2.393 & .017 & * \\
\texttt{Phonological $\times$ Women} & -0.274 & 0.259 & -1.055 & .292 & \\
\bottomrule
\label{tab:stats}
\end{tabular}
\vspace{0.5em}

\raggedright
\textit{Note.} Significance codes: *** $p < .001$, ** $p < .01$, * $p < .05$, \textbf{.} $p < .1$.\\
Model: Negative Binomial GLMM with random intercept for speaker. ``Morphosyntax'' and ``men'' were used as reference levels.
\end{table}

\begin{figure}[H]
\begin{flushleft}
    \centering
    \includegraphics[width=0.8\textwidth]{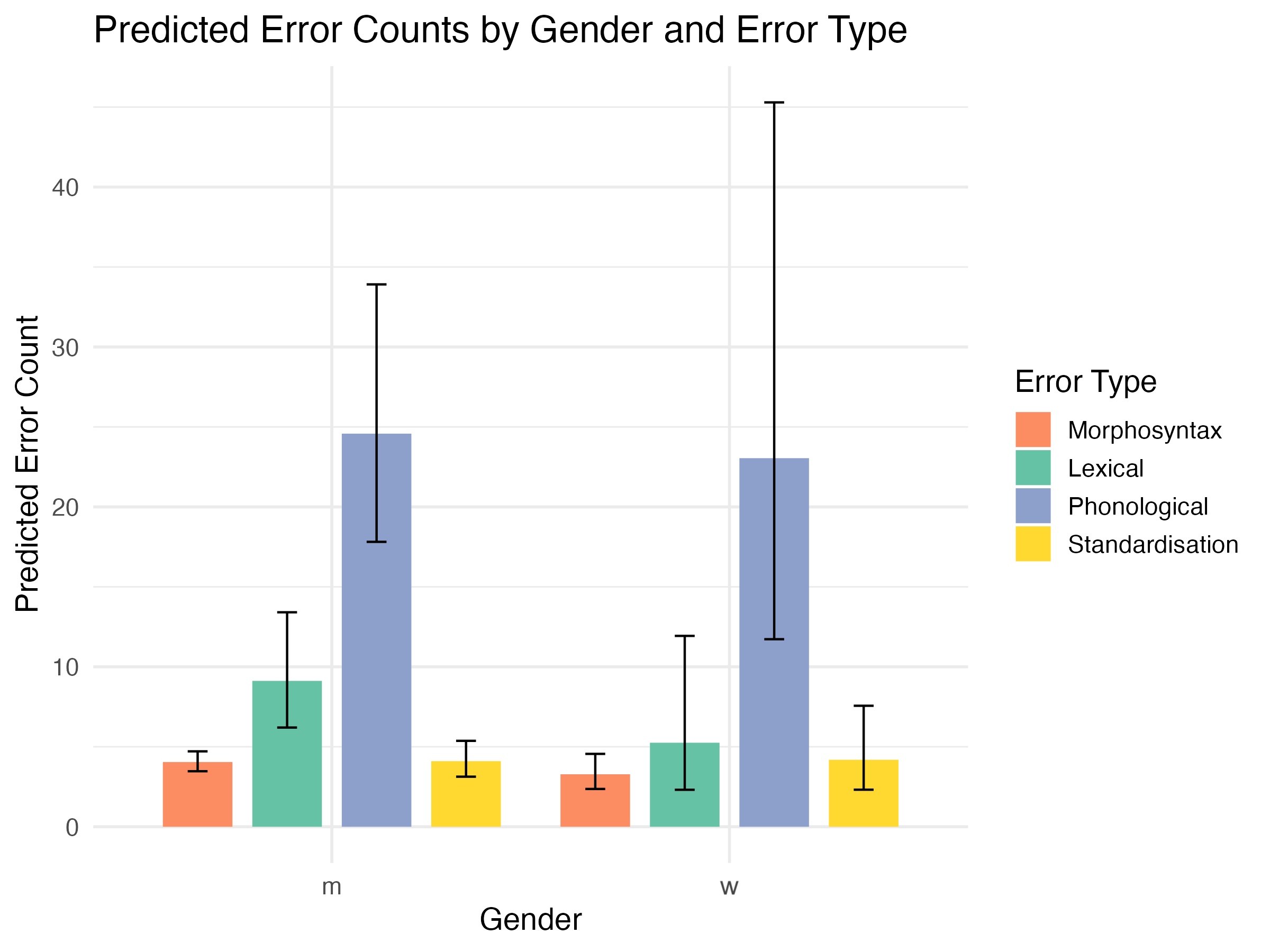}
    \caption{Predicted error counts by gender and error type.}
    \label{fig:preds_gender}
\end{flushleft}
\end{figure}

While the main effect of gender was not statistically significant ($p = .188$), the Standardisation × Women interaction was significant ($p = .017$), indicating that the gender effect for standardisation differed significantly from that of the reference error category (morphosyntax). This interaction should not be interpreted as evidence of a significant gender difference within standardisation errors themselves; post-hoc comparisons showed no such difference. To understand the interaction, Figure~\ref{fig:preds_gender} shows predicted error counts by gender and error type. Across both genders, phonological errors were predicted to be the most frequent by a wide margin (men: $\hat{\mu} = 24.58$, 95\% CI [17.81, 33.91]; women: $\hat{\mu} = 23.05$, 95\% CI [11.73, 45.29]). In contrast, standardisation (men: $\hat{\mu} = 4.10$; women: $\hat{\mu} = 4.19$) and morphosyntactic errors (men: $\hat{\mu} = 4.05$; women: $\hat{\mu} = 3.28$) were predicted to occur substantially less frequently. The interaction between gender and error type is visually apparent primarily in the lexical category, where men are predicted to produce more lexical errors ($\hat{\mu} = 9.12$, 95\% CI [6.20, 13.41]) than women ($\hat{\mu} = 5.25$, 95\% CI [2.31, 11.93]).

Pairwise comparisons of estimated marginal means \citep{emmeans}, using Tukey-adjusted \textit{p}-values for multiple comparisons, revealed that phonological errors were significantly more frequent than all other error types. Specifically, phonological errors were over five times more likely than standardisation errors (ratio $\approx$ 5.75, \textit{p}~<~.0001), and more than three times more likely than lexical errors (ratio $\approx$ 3.43, \textit{p}~<~.0001). Lexical errors also occurred significantly more frequently than morphosyntactic errors (ratio $\approx$ 1.90, \textit{p}~<~.0001), and were 1.67 times more likely than standardisation errors (\textit{p} = .0002). In contrast, morphosyntactic and standardisation errors occurred at similar rates (ratio $\approx$ 0.88, \textit{p} = .7881), with no significant difference. Pairwise comparisons showed a significant gender difference in lexical errors, with the ASR producing more lexical errors for men’s than women’s speech ($\hat{\beta} = 0.55$, p = .008). No significant gender differences were observed for phonological, morphosyntactic, or standardisation errors (p > .38 for all).

\subsubsection{Phonological errors}
\label{subsubsec4.1.3}

Having established the overall distributional patterns in error counts, we now turn to a more fine-grained linguistic analysis of the individual error types, beginning with phonological errors. Phonological features of Newcastle English have presented the most challenges to the ASR system by far, which is not a surprise. The local dialect boasts numerous salient features that are uncommon not just in SSBE but also in the rest of the North of England. 

We break down the phonological errors first by a subcategory -- this distribution is given in Table~\ref{tab:phon_errors}. The groups here are quite unequal, as we classified all errors related to vowel quality under one sub-category before breaking them down further. At the same time, we have included such specific features as `g'-dropping or the clear /l/, since they are stand-alone features and quite salient in Newcastle. Another reason was that at this stage of the analysis, we wanted to keep the number of sub-categories manageable.

The following passages explain the three leading error subcategories in more detail.

\begin{table}[H]
\centering
\caption{Frequency distribution of phonological errors by subcategory. Percentages indicate the share of total phonological errors.}
\begin{tabular}{p{5cm} p{2cm} p{2cm}}
\toprule
\textbf{Error Subcategory} & \textbf{Count} & \textbf{Percentage} \\
\midrule
Vowel quality                 & 721 & 39.5\% \\
Glottalisation/glottal stop               & 414 & 22.7\% \\
Monophthongisation    & 403 & 22.1\% \\
G-dropping            & 171 & 9.4\%  \\
Clear /l/             & 45  & 2.5\%  \\
Phonetic reduction    & 23  & 1.3\%  \\
H-dropping            & 19  & 1.0\%  \\
T-to-R                & 15  & 0.8\%  \\
Consonant             & 9   & 0.5\%  \\
Aspiration            & 3   & 0.2\%  \\
HappY tensing         & 3   & 0.2\%  \\
\bottomrule
\end{tabular}
\label{tab:phon_errors}
\end{table}

\paragraph{Vowel Quality}
\label{paragraph4.1.3.1}

\begin{figure}[H]
\begin{flushleft}
    \centering
    \includegraphics[width=1\textwidth]{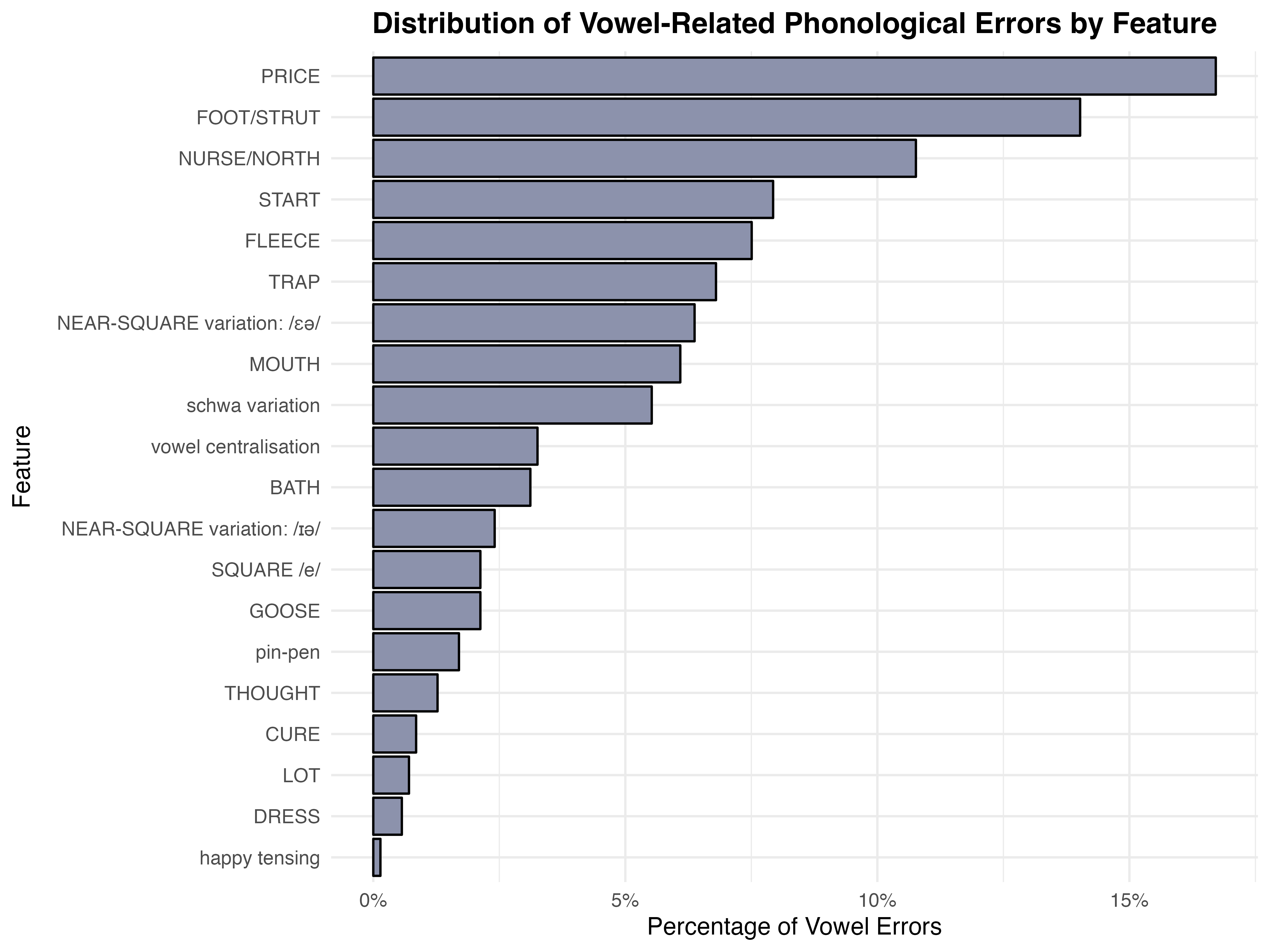}
    \caption{Distribution of vowel-related phonological errors by feature (excluding monophthongisation).}
    \label{fig:phon_by_feature}
\end{flushleft}
\end{figure}

We continue by breaking down the errors further and classifying them according to the specific phonological feature identified as the likely source of the ASR misrecognition based on manual analysis of the speech and comparison with the reference transcription. Figure~\ref{fig:phon_by_feature} lists all phonological errors related to vowel quality. 

Errors related to the \textsc{price} vowel were the most common error type. The \textsc{price} vowel can be realised as \textipa{[EI]}, \textipa{[i:]}, or \textipa{[A:]} in Newcastle. This is reflected in the kinds of errors ASR produced, e.g., ``reminded'' → ``remained'', ``like'' → ``lake'', ``my'' → ``may'' (all pronounced with \textipa{[EI]}); ``mind'' → ``mean'', ``clientele'' → ``clean tell'' (pronounced with \textipa{[i:]}); ``wife's'' → ``was'', ``bikes'' → ``backs'' (pronounced with \textipa{[A:]}).

Two mergers -- \textsc{foot/strut} and \textsc{nurse/north} -- were the second and third leading causes of error. Since we expect both vowels in \textsc{foot/strut} to be realised as \textipa{[U]}, the root of errors is again evident. Some examples include ``son'' → ``soon'', ``gun'' → ``good''. In other cases, the vowel is not immediately replaced by an SSBE equivalent, and a less phonetically close substitution is used: ``fun'' → ``phone'', ``cut'' → ``could'', ``other'' → ``either''. A similar trend can be seen in the \textsc{nurse/north} merger, where a more rounded \textsc{nurse} vowel is expected. Substitutions include ``urban'' → ``open'', ``work'' → ``wake'', ``serving'' → ``saving''. The reverse when the \textsc{north} vowel was affected was less frequent but also possible, e.g. ``courted'' → ``quoted'', ``for'' → ``first''.

Interestingly, the \textsc{start} vowel has also received many errors, even though it is not considered as salient in Newcastle as e.g., the \textsc{price} or \textsc{nurse/north} vowels. Errors include ``part'' → ``pot'', ``dark'' → ``dog'', ``fart'' → ``fort''. Generally, substitutions with the [o] or [\textopeno:] vowels were very common. This pattern is consistent with the Newcastle realisation of the \textsc{start} vowel, which can range between [\textscripta:] and [\textturnscripta:] \citep{pearce2009perceptual}. The more rounded [\textturnscripta:]-like realisations may therefore contribute to substitutions involving [o]- or [\textopeno:]-like vowels. It could also be that the quality here is similar to the \textsc{bath} vowel where no retraction is expected. In \textsc{bath}, we found such errors as ``France'' → ``front'', ``laugh'' → ``left'', ``aghast'' → ``a guest'', ``baths'' → ``bus'' (a repeated error).

The \textsc{fleece} vowel constitutes another prominent sub-category of vowel-related errors. It is logical to discuss it in combination with the \textsc{goose} vowel since both can be realised closer to cardinal vowels in Newcastle English. One of the most common error patterns with \textsc{fleece} was a substitution akin to ``see'' → ``say'' (that particular error alone occurred 17 times) or ``keys'' → ``case''. Other kinds of substitutions like ``we'' → ``well'', ``clean'' → ``claim'', ``sheep'' → ``ship'' or even ``being'' → ``buying'' occurred as well. Within the \textsc{goose} vowel, ``blooming'' → ``plumbing'', ``snooker'' → ``snugger'' and ``snooker'' → ``soccer'', as well as ``goods'' → ``codes'' were some of the errors.

As for the other, less numerous errors, the \textsc{near/square} variation also warrants further discussion. As Figure \ref{fig:phon_by_feature} shows, we have divided it into two types depending on the realisation: /\textepsilon \textschwa/ or /\textsci \textschwa/. Examples of the \textsc{near/square} variation with /\textepsilon \textschwa/ include ``pair'' → ``pay'', ``forty year'' → ``forty yeah'', ``shared'' → ``shed''. As for the \textsc{near/square} variation with /\textsci \textschwa/, errors included ``beer'' → ``bay'', ``realise'' → ``relies'', ``weird'' → ``we had''. Significantly less frequent is the \textsc{square} vowel pronounced with an [e] and resulting in errors such as ``air'' → ``eh'', ``their'' → ``the''. Since the quality of the vowel is so distinct, we included it as a separate feature and therefore propose to treat it as a separate cause of errors.

Finally, there are the remaining vowel features in Figure \ref{fig:phon_by_feature} that are not particularly salient in Newcastle but still caused ASR errors. We include them because their realisation differs from what would be expected in SSBE, which gives us reasons to believe that their non-standard nature causes ASR to fail to recognise these tokens. For instance, we noted several errors related to the \textsc{pin}/\textsc{pen} merger, such as ``will'' → ``well'', ``tent'' → ``tint''. The \textsc{trap} vowel pronounced as [a] \citep{beal2004english, hickey2015north} was another common error source, e.g. ``blanked'' → ``blunt'', ``dragged'' → ``drugged'', ``bad'' → ``but''. Overall, however, these errors are far less numerous, making it clear that particularly salient regional vowel features cause the most difficulties for ASR.

\paragraph{Near-glottalisation}
\label{paragraph4.1.3.2}

Near-glottalisation of /p,t,k/ was the second leading cause of error in phonological errors. Within this category, /t/-glottalisation caused 13.9\% of all phonological errors and 61.4\% of glottalisation errors in particular (n=254). Examples include ``forgot to tell'' → ``forgot tell'', ``satan'' → ``saying'', ``bottom'' → ``bump'', ``convicts'' → ``comics'', ``saturday'' → ``sunday''. Next, /k/-glottalisation (4\% of all phonological errors, n=73) caused errors such as ``blanked'' → ``blunt'', ``every waking minute'' → ``every week a minute'' and several cases of ``I can remember'' → ``I remember''. Lastly, /p/-glottalisation (2.2\% of all phonological errors, n=40) resulted in errors such as ``skipping'' → ``skiing'', ``supermarket'' → ``a market'', ``compensation'' → ``conversation''. Glottal stops were included in this category as well, causing an additional 2\% of all phonological errors (e.g., ``spat'' → ``spa'').

\paragraph{Monophthongisation}
\label{paragraph4.1.3.3}

\textsc{face}- and \textsc{goat}-monophthongisation caused a notable proportion of phonological errors (15.2\% and 7.5\% of all phonological errors, respectively). \textsc{face}-monophthongisation errors included ``failed'' → ``field'', ``fail'' → ``feel'', ``saying'' → ``seeing'' (or ``say'' → ``see'', n=36), ``pay'' → ``peer'', ``sale'' → ``seal''. \textsc{goat}-monophthongisation caused errors as well, e.g., ``loathe'' → ``love'', ``tone'' → ``tune'', ``coat'' → ``called'', ``ropes'' → ``robs'', ``Rome'' → ``room''. 

\subsubsection{Morphosyntactic errors}
\label{subsubsec4.1.4}

Morphosyntactic errors accounted for 7.8\% of all errors (n=235), making it one of the less prominent errors types. However, the errors in this subcategory are representative of the challenges that dialectal features cause to ASR systems. Table ~\ref{tab:syntax_errors} demonstrates that almost all errors are related to regional variation in verb usage (47.4\%, n=111) or local pronoun usage (43.6\%, n=102).

\begin{table}[H]
\centering
\caption{Frequency distribution of morphosyntactic errors by subcategory. Percentages indicate the share of total morphosyntactic errors.}
\begin{tabular}{p{4cm} p{2cm} p{2cm}}
\toprule
\textbf{Error Subcategory} & \textbf{Count} & \textbf{Percentage} \\
\midrule
verb paradigm   & 111 & 47.4\% \\
pronoun         & 102 & 43.6\% \\
tenses          & 17  & 7.3\%  \\
plurals         & 4  & 1.7\%  \\
\bottomrule
\end{tabular}
\label{tab:syntax_errors}
\end{table}

\begin{figure}[H]
\begin{flushleft}
    \centering
    \includegraphics[width=1\textwidth]{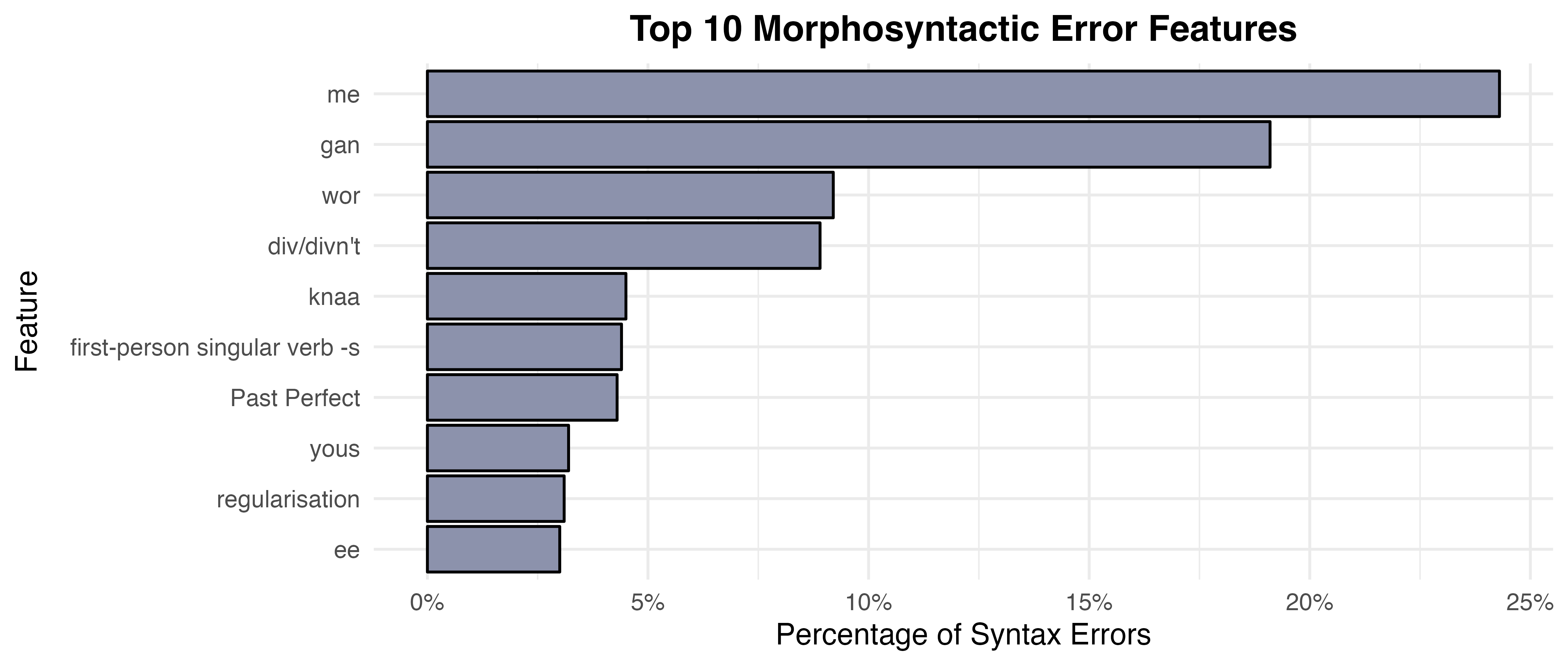}
    \caption{Distribution of the most common morphosyntactic errors by feature.}
    \label{fig:top10_syntax}
\end{flushleft}
\end{figure}

We elaborate on these four subcategories in the following sections. For a general overview, Figure \ref{fig:top10_syntax} also shows 10 most frequent morphosyntactic features that caused errors in this category -- most of them pronouns and verbs.

\paragraph{Verb paradigm}
\label{paragraph4.1.4.1}

Local variations of verbs caused the most morphosyntactic errors in our dataset. Among these, ``gan'', ``div/divn't'', and ``knaa'' were the most common. First person singular -s in verbs was another major source of errors.

The verb ``gan'' (meaning ``go'') was consistently mistranscribed, phonetic similarity being seemingly the decisive factor in ASR output. Examples include ``gan'' → ``gun'', ``gan'' → ``gone'', ``gan'' → ``can''. Similarly, ``div/divn't'' (regional forms of ``do'' and ``don’t'') were transcribed as ``did/didn't'' (as well as sometimes ``don't'', which is semantically correct and can also be argued to be standardisation). ``Knaa'' (meaning ``know'') was sometimes standardised as ``know'', but forms like ``yknaa'' were consistently mistranscribed as ``yeah'' or deleted from the transcript completely. First-person singular verb -s in, e.g., ``I says'' was repeatedly mistranscribed as ``I said'', which distorts the meaning.

\paragraph{Pronouns}
\label{paragraph4.1.4.2}

\citet{serditova25_interspeech} demonstrated that local pronouns ``wor'' and ``yous'' cause significant difficulties for ASR, with ``wor'' also showing the impact of age on the number of errors speakers receive. In the expanded analysis, the two pronouns were the second and third most common cause of all pronoun errors (18.9\% and 11.7\% of all pronoun errors, respectively). In our previous study, we did not have the opportunity to elaborate on these errors qualitatively and provide examples. Furthermore, we need to discuss the leading cause of all morphosyntactic errors (see Figure \ref{fig:top10_syntax}) -- the local possessive pronoun ``me'' used instead of ``my''.

Errors connected to the local use of the pronoun ``me'' fall both into the category ``Morphosyntax'' and ``Standardisation''. For the second type, see Section \ref{subsubsec4.1.6} where we demonstrate that the meaning is kept intact despite these errors and only their regional realisation suffers. The ``me'' errors that we classified as morphosyntactic errors are the ones where the ground truth meaning is completely or partially distorted. Examples include ``me own'' → ``New Orleans'', ``got me discharge'' → ``got me discharged'', ``me night'' → ``midnight''.

Errors related to the local pronoun ``wor'' also fell under both ``Morphosyntax'' and ``Standardisation'' depending on the output. If ``wor'' was transcribed as ``our'', we counted it as a standardisation error because the meaning did not suffer. However, in many cases ``wor'' was transcribed in a way that we can not explain semantically, e.g. ``wor'' → ``a'', ``wor'' → ``out''. In several cases, the pronoun was simply omitted from the transcript. In other cases, a phonetically close transcription was used: ``wor'' → ``were'', ``wor'' → ``for'', ``wor'' → ``where''. Finally, the related pronoun ``worselves'' (meaning ``ourselves'') caused difficulties too and was transcribed as ``what''.

Many instances of the pronoun ``yous'' were also classified as standardisation because the ASR output was ``you'', which is not semantically incorrect. However, it is most likely that the decision was taken based on phonetic similarity, so the acoustic model deemed it the most plausible option. We are not able to check whether the pronoun ``yous'' in included in Rev AI's vocabulary. We classified several ``yous''-related errors as morphosyntactic errors, e.g. ``yous'' → ``I've'', ``yous'' → ``news'', ``if yous'' → ``few'', all of which also point to phonetic similarity being the most plausible decisive factor.

As for the rest of the pronoun-related errors, there were several instances when the pronoun ``us'' (to mean ``me'') was mistranscribed, e.g. ``start us off in me career'' → ``start as off in me career''. The pronoun ``meself'' was consistently mistranscribed as ``said'' or even ``Michelle''. The local realisation ``ee'' (meaning ``you'') resulted in deletion errors. The local realisation ``theirself'' was mistranscribed as ``myself''.

\paragraph{Tenses and plurals}
\label{paragraph4.1.4.3}

In a few instances, tenses were mistranscribed, with most errors connected to Present Perfect or Past Perfect, e.g., ``had happened'' → ``has happened'', ``we'd done'' → ``we've done''. The cases where plural nouns were used without the -s ending were commonly standardised to include the -s and therefore fall under standardisation errors (e.g., ``four year'' → ``four years''). There were additional instances of specific errors like ``many a time'' → ``many times'' or a case of deictic syntax ``it’s an awkward one this'' → ``it’s an awkward one there'' that did not contribute majorly to the overall error count and will not be discussed further.

\subsubsection{Lexical errors}
\label{subsubsec4.1.5}

\begin{figure}[H]
\begin{flushleft}
    \centering
    \includegraphics[width=0.8\textwidth]{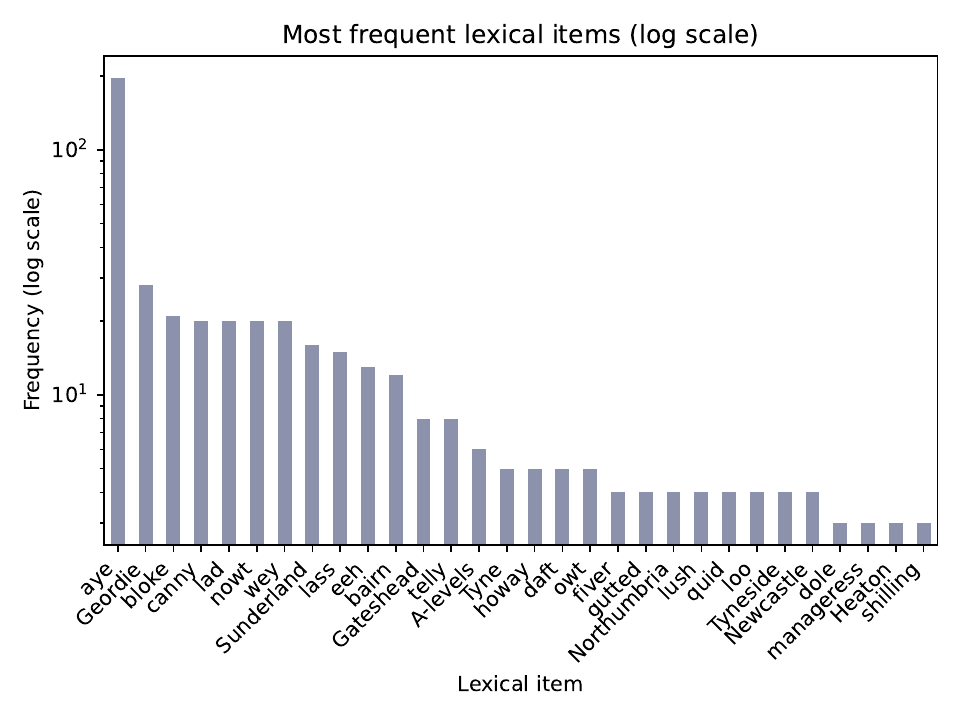}
    \caption{Most frequent lexical items causing ASR errors, log-scaled.}
    \label{fig:top_lexical}
\end{flushleft}
\end{figure}

Lexical errors constituted nearly 20\% of the total error count and are the second most prominent error type after phonological errors. When classifying errors, we divided lexical errors into two major sub-categories: toponyms and vocabulary errors. The most frequent lexical items are given in Figure \ref{fig:top_lexical}.

Toponyms accounted for 13.2\% of all lexical errors in the dataset. Out of these, such place names as Sunderland (20.5\% of all toponyms errors), Gateshead (10.3\%), Tyne (6.4\%), Newcastle (5.1\%), Northumbria (5.1\%), and Tyneside (5.1\%) were the most common occurrences. Other place names like Cullercoats, North Shields, South Shields, and Toon resulted in errors as well. Since we used commercial ASR for this study, it is not feasible to determine whether these toponyms are simply absent from the dictionary or there is a different root cause of these errors.

The rest of the lexical errors were categorised under ``Vocabulary''. We believe they were caused by local lexical items that might indeed not be included in standard training models due to their both regional and sometimes also colloquial nature. One lexical item stands out in frequency: the item ``aye'', commonly used in Newcastle and the broader North East, and the most frequent cause of lexical errors for both men and women. It is a non-standard affirmative lexical item, which is used in place of ``yes'' and is considered a regionally marked lexical feature of Geordie English. Taking up 38.3\% of all local vocabulary errors, it is a salient feature of Newcastle English consistently mistranscribed by ASR.

Other local vocabulary items that caused issues were ``Geordie'' (5.5\% of all local vocabulary errors), ``bloke'' (4.1\%), ``canny'' (meaning ``good'', 3.9\%), ``lad'' (3.9\%), ``nowt'' (meaning ``nothing'', 3.9\%), ``wey'' (meaning ``why yes'', an affirmation, 3.9\%) ``lass'' (a woman or girl, 2.9\%), ``bairn'' (a child, 2.3\%). These local lexical items are strong markers of regional identity and cultural belonging. The high rate of ASR errors suggests a bias against dialectal forms. Such systematic misrecognition risks marginalising local speech patterns and undermines the linguistic legitimacy of regional speakers.

Other lexical items which were used less frequently and therefore received fewer errors but are still important to mention include ``lush'' (meaning ``very good''), ``owt'' (``anything''), ``geet'' (``great''). Surprisingly, more widespread lexical items like ``A-levels'', ``loo'', and ``telly'' (see also Section \ref{subsubsec4.1.6}) received errors as well, indicating that the ASR system struggles not only with regionalisms but also with commonly used British English terms.

\begin{figure}[H]
\begin{flushleft}
    \centering
    \includegraphics[width=1.2\textwidth]{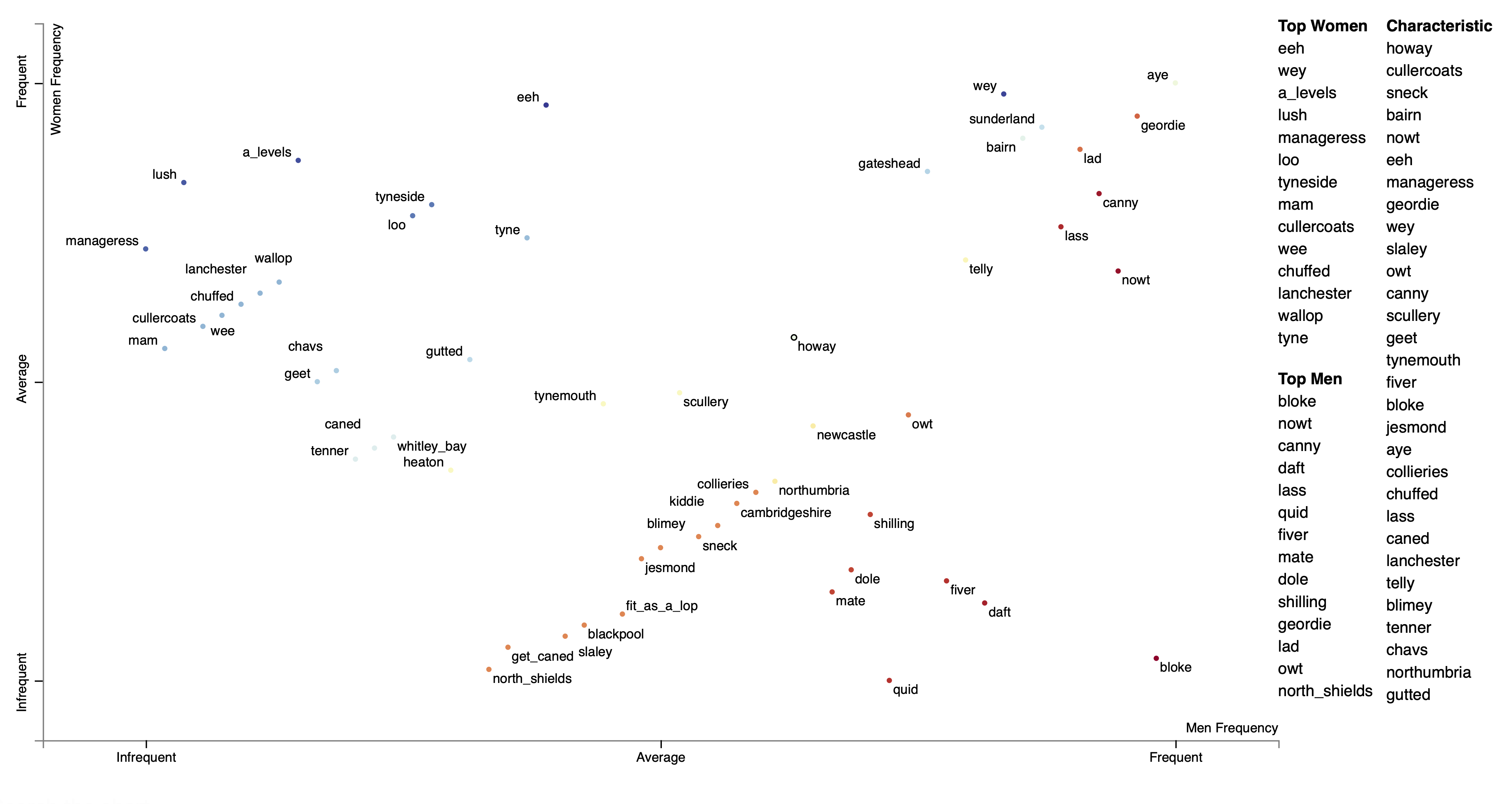}
    \caption{Most frequent lexical errors, distributed by gender. Plotted using \texttt{scaled F score} metric using \texttt{scattertext} \citep{kessler2017scattertext}. An interactive version of this figure can be found in the OSF repository.}
    \label{fig:scattergender}
\end{flushleft}
\end{figure}

To examine which lexical items contribute to the gender difference in lexical errors identified in the statistical analysis, Figure \ref{fig:scattergender} compares the distribution of problematic lexical items across men's and women’s speech. The x-axis shows frequency rates for men, and the y-axis for women. The items that are particularly frequent for men, including ``bloke'', ``nowt'', ``canny'', are all representative of the local vocabulary. For women, the most frequent items include words that are dialectal for the UK at large and not just Newcastle, such as ``A-levels'' or ``loo''. Interestingly, women received notably more errors in toponyms. Furthermore, items such as ``aye'', ``Geordie'', ``lad'', ``Sunderland'' were common sources of error for both genders.  

\subsubsection{Standardisation and spelling errors}
\label{subsubsec4.1.6}

In this section, we turn to two other sets of errors: standardisation and spelling. Standardisation errors refer to instances where dialectal features were replaced with forms associated with Standard Southern British English (e.g., ``me life'' → ``my life'', ``wor'' → ``our''). Spelling errors are somewhat similar in nature -- they mostly involved substitutions of British spellings with their American counterparts. This pattern reflects the strong influence of American English in ASR training data, leading systems to favour American orthographic conventions over British ones even when it is indicated that the audio file is in UK English (which Rev AI allows to do before processing the file).

The majority of standardisation errors involved pronouns (56.5\%), where dialect-specific forms were replaced with standard equivalents. Conjunctions (16.9\%, e.g., ``cause'' → ``because'') and vocabulary items (13.7\%, e.g., ``tummy'' → ``stomach'') were also frequent targets of correction. Interestingly, while in other instances the word ``telly'' was mistranscribed completely (see Section  \ref{subsubsec4.1.5}), we found one instance of ``telly'' → ``television''. The standardisation ``round'' → ``around'' was another frequent occurrence.

Less common were changes to plurals (e.g., ``five year\_ ago'' → ``fives years ago'', ``ninety six pound\_'' → ``96 pounds''), verb paradigms (e.g., ``knaa'' → ``know''), and other grammatical categories such as tense and adverbs, reflecting a tendency to standardise non-standard grammatical features. Conflation of past tenses has proven to be problematic in this instance too, with such corrections as ``haven't give'' → ``haven't given'', ``I've never really spoke'' → ``I've never really spoken''.

Spelling errors were not numerous but are still worth mentioning. Some notable instances were: ``license'' → ``licence'', ``practise'' → ``practice'', ``learnt'' → ``learned'', ``burnt'' → ``burned'', ``maths'' → ``math'', ``spoilt'' → ``spoiled'', ``favourite'' → ``favorite''. These substitutions reflect American rather than British spelling conventions — an issue that would not be restricted to Newcastle speakers but affect the British population at large. It is particularly concerning that such substitutions occur even when UK English is specified prior to processing, suggesting a systemic bias in the language models underpinning ASR systems.

\subsection{Acoustic analyses}
\label{subsec:acousticanalyses}

The analyses presented above draw on expert auditory evaluation conducted by two trained sociophoneticians and provide detailed insights into the linguistic patterns underlying ASR performance. To further substantiate these findings, we complement the auditory analysis with a focused case study in which we acoustically assess the degree of \textsc{face}- and \textsc{goat}-monophthongisation and centering in a selected group of speakers. This allows us to examine whether speakers with more \textsc{face}- and \textsc{goat}-related ASR errors show a greater tendency toward monophthongal or centering realisations of the corresponding diphthongs, which are established regional variants in Newcastle (see Table \ref{tab:tyneside_features_phon}). In doing so, we integrate qualitative and acoustic evidence to more systematically account for the relationship between phonetic variation and ASR errors. This approach builds on our earlier work on a syntactic feature (local pronouns; \citet{serditova25_interspeech}), where we similarly demonstrated that local realisations have a measurable impact on ASR error counts.

\begin{figure}[H]
\begin{flushleft}
    \centering
    \includegraphics[width=1\textwidth]{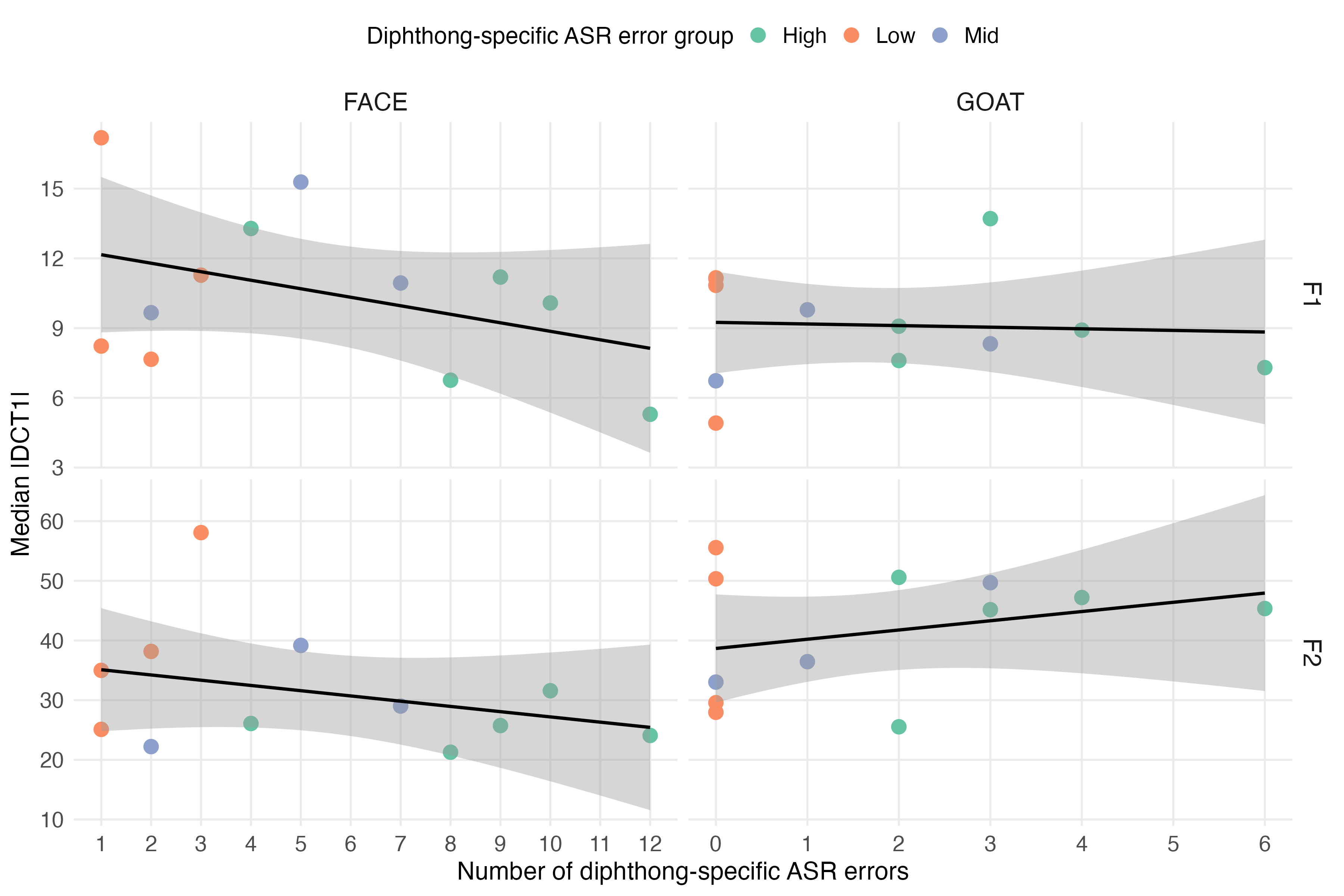}
    \caption{Relationship between diphthong-specific ASR error counts and speakers'
median absolute DCT1 values for F1 and F2 in \textsc{face} and \textsc{goat}.
Points represent individual speakers, coloured by overall ASR error group;
black lines show descriptive linear trends with 95\% confidence intervals.}
    \label{fig:global_DCT1}
\end{flushleft}
\end{figure}

We first assessed this relationship globally across the 12 speakers before
examining individual speakers' formant trajectories in greater detail. For
each speaker, we calculated the median absolute DCT1 value for F1 and F2
across their \textsc{face} and \textsc{goat} tokens. These values were
compared with the corresponding vowel-specific ASR error counts using
Spearman's rank correlations. $|\mathrm{DCT1}|$ captures the magnitude of
change in the formant trajectory; consequently, lower absolute DCT1 values
indicate less formant movement and, for diphthongal vowels, a more
monophthongal realisation \citep{watson1999acoustic, elvin2016dynamic}. Figure~\ref{fig:global_DCT1} shows the relationship between vowel-related ASR error counts and speakers' median absolute DCT1 values for F1 and F2, separately for \textsc{face} and \textsc{goat}\footnote{The Low, Mid, and High categories refer to speakers' frequency of both \textsc{face}- and \textsc{goat}-related ASR errors. A speaker may belong to the Mid-range or High group while having few or no errors for one of the two vowels. The diphthong-specific error counts are used separately in the analyses presented below.}. At the speaker level, \textsc{face} showed weak negative correlations
between ASR error count and median absolute DCT1 for both F1
($\rho=-.29$, $p=.365$) and F2 ($\rho=-.24$, $p=.462$).
For \textsc{goat}, the correlations were also weak, with a negative
association for F1 ($\rho=-.10$, $p=.753$) and a positive association
for F2 ($\rho=.15$, $p=.644$). None of these correlations reached
statistical significance. Given the small speaker sample ($n=12$),
these analyses have limited power to detect speaker-level associations
and should therefore be interpreted as exploratory.

Having explored the global pattern, we now turn to a selected set of figures representative of Low-error, High-error, and Mid-range error speakers\footnote{While we have tried to keep the x- and y-axes as uniform as possible, the F1 axis varies slightly for a more close-up and precise visualisation. For the rest of the individual speaker plots, see ~\ref{appendix}.}. Figure \ref{fig:decten2y10i026} demonstrates the formant trajectories for the two diphthongs of an older women who received the least number of monophthongisation errors (n=1). Evidently, the glides for both \textsc{face} and \textsc{goat} are quite pronounced. Their direction is also what would be expected in a more mainstream realisation, and they do not have the centering quality that these diphthongs in Newcastle tend to have.

\begin{figure}[H]
\begin{flushleft}
    \centering
    \includegraphics[width=0.6\textwidth]{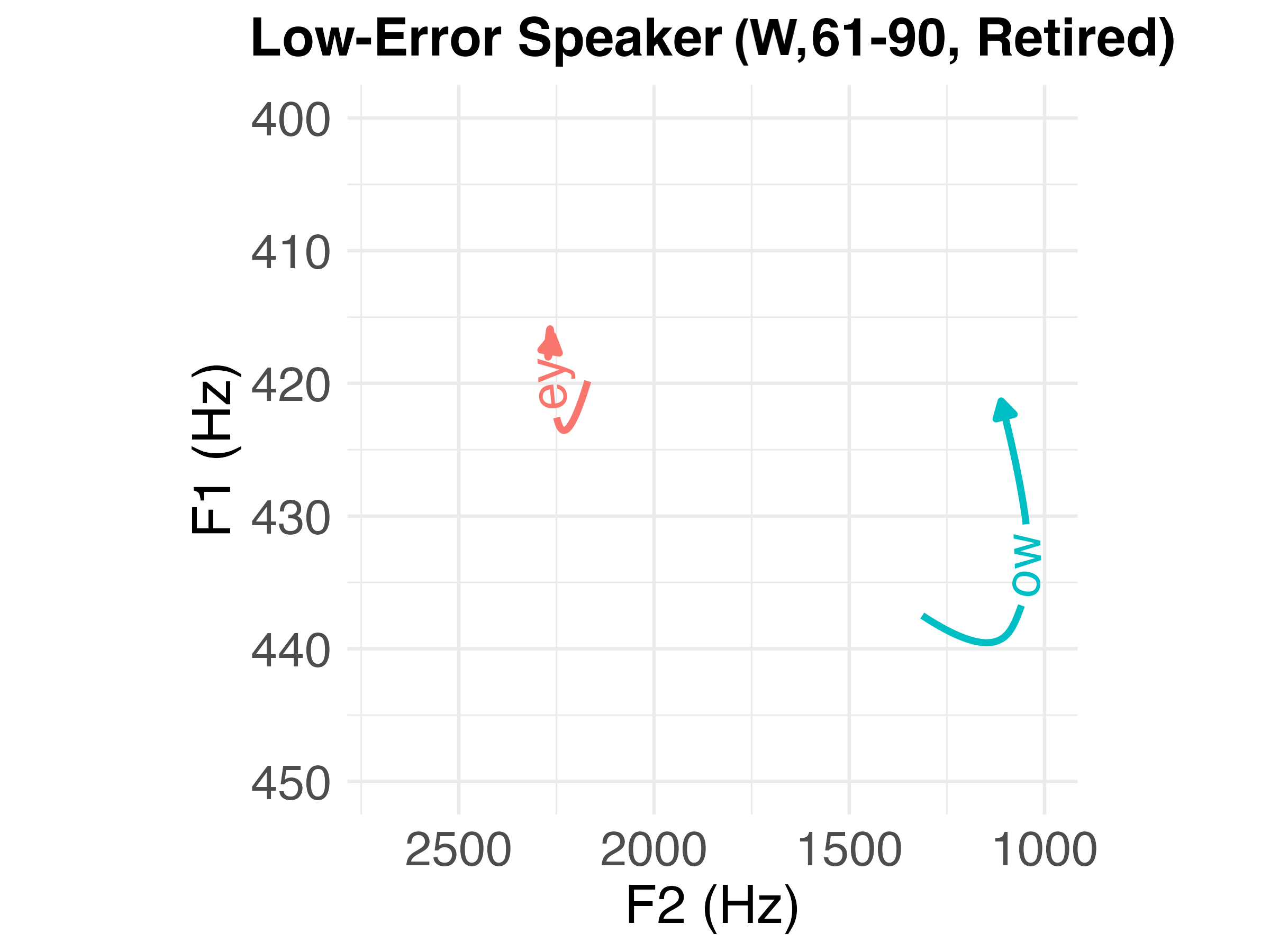}
    \caption{Formant trajectory of a speaker with the least number of errors.}
    \label{fig:decten2y10i026}
\end{flushleft}
\end{figure}

\begin{figure}[H]
\centering
\begin{subfigure}{0.52\textwidth}
    \centering
    \includegraphics[width=\linewidth]{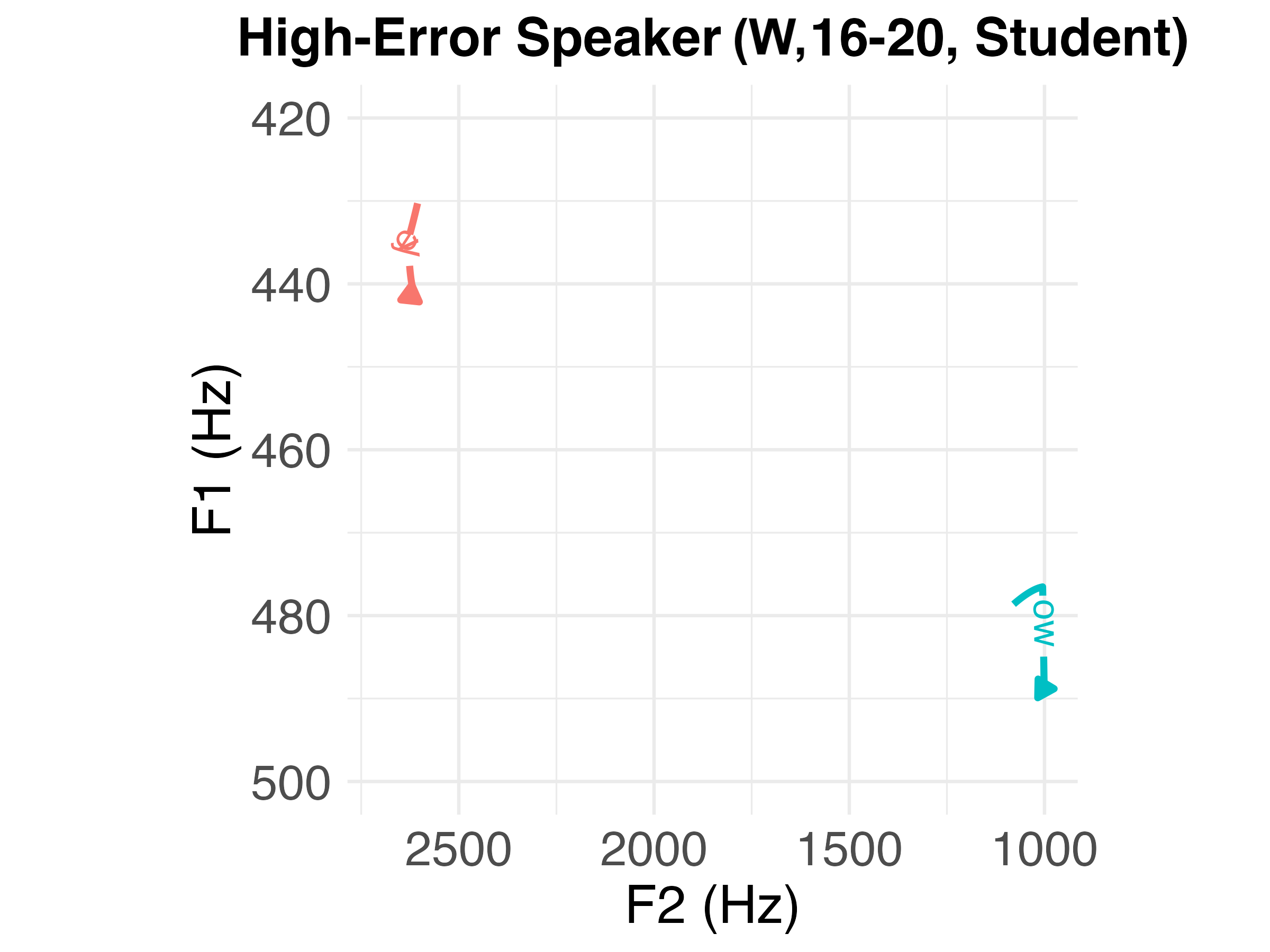}
\end{subfigure}%
\begin{subfigure}{0.52\textwidth}
    \centering
    \includegraphics[width=\linewidth]{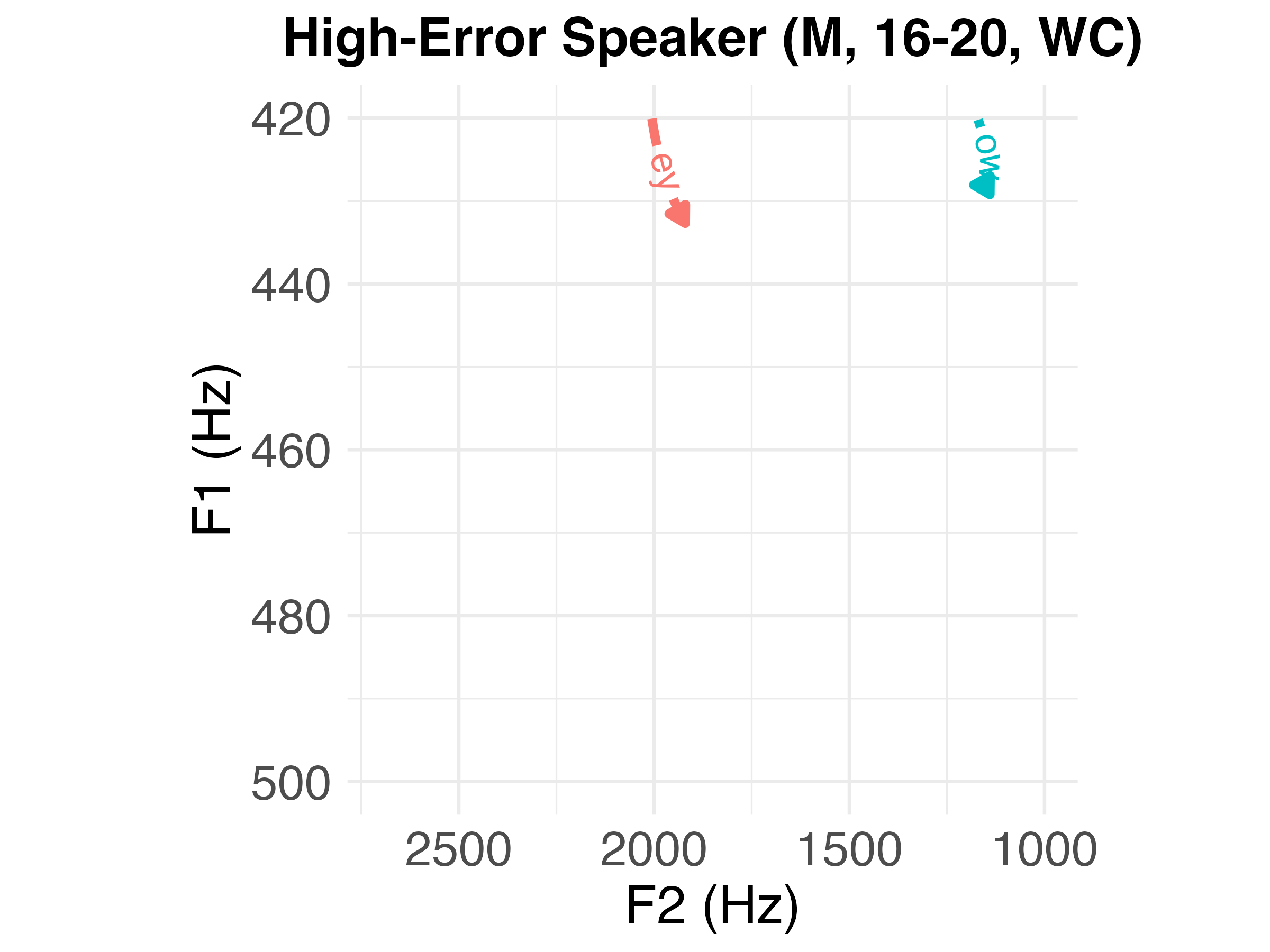}
\end{subfigure}
\caption{Formant trajectories of speakers with the highest number of errors.}
\label{fig:JessicaTimothy}
\end{figure}

Contrary to that, Figure \ref{fig:JessicaTimothy} 
shows two speakers of the same age (from the same audio file) who received the highest number of errors per file $(n_{\text{men}} = 15;\ n_{\text{woman}} = 12)$, distributed rather equally. The glides here are downward on the F1 trajectory, demonstrating the centering quality of the regional realisations \textipa{[i@]} and \textipa{[u@]}. Interestingly, the woman's \textsc{goat} vowel is also noticeably lower than that of the man. In the case of the man, the glide in \textsc{goat} is barely noticeable, making the realisations akin to \textsc{goat} [\textbaro] a plausible explanation of why ASR struggled to correctly identify the vowel.

\begin{figure}[H]
\centering
\begin{subfigure}{0.52\textwidth}
    \centering
    \includegraphics[width=\linewidth]{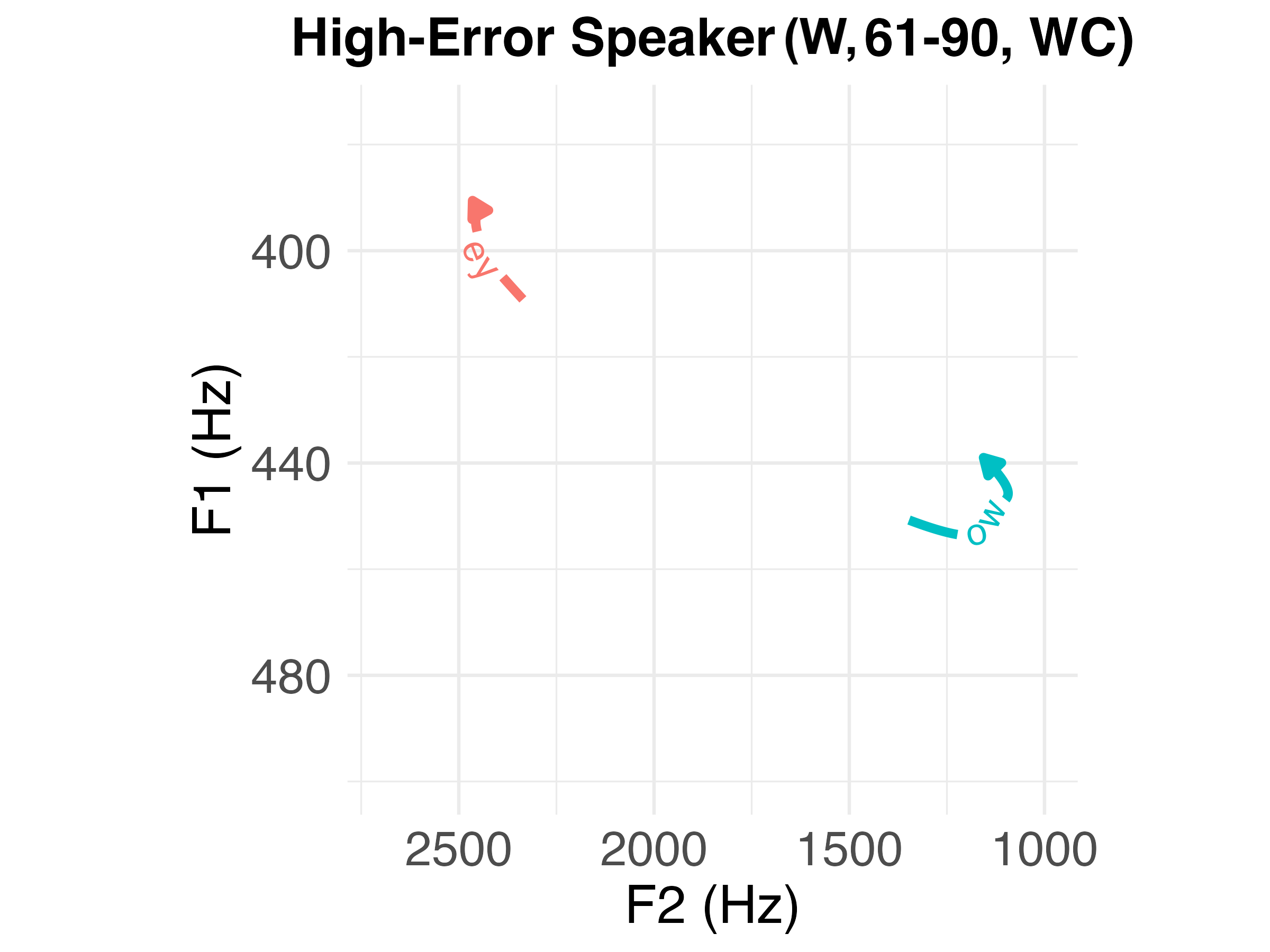}
\end{subfigure}%
\begin{subfigure}{0.52\textwidth}
    \centering
    \includegraphics[width=\linewidth]{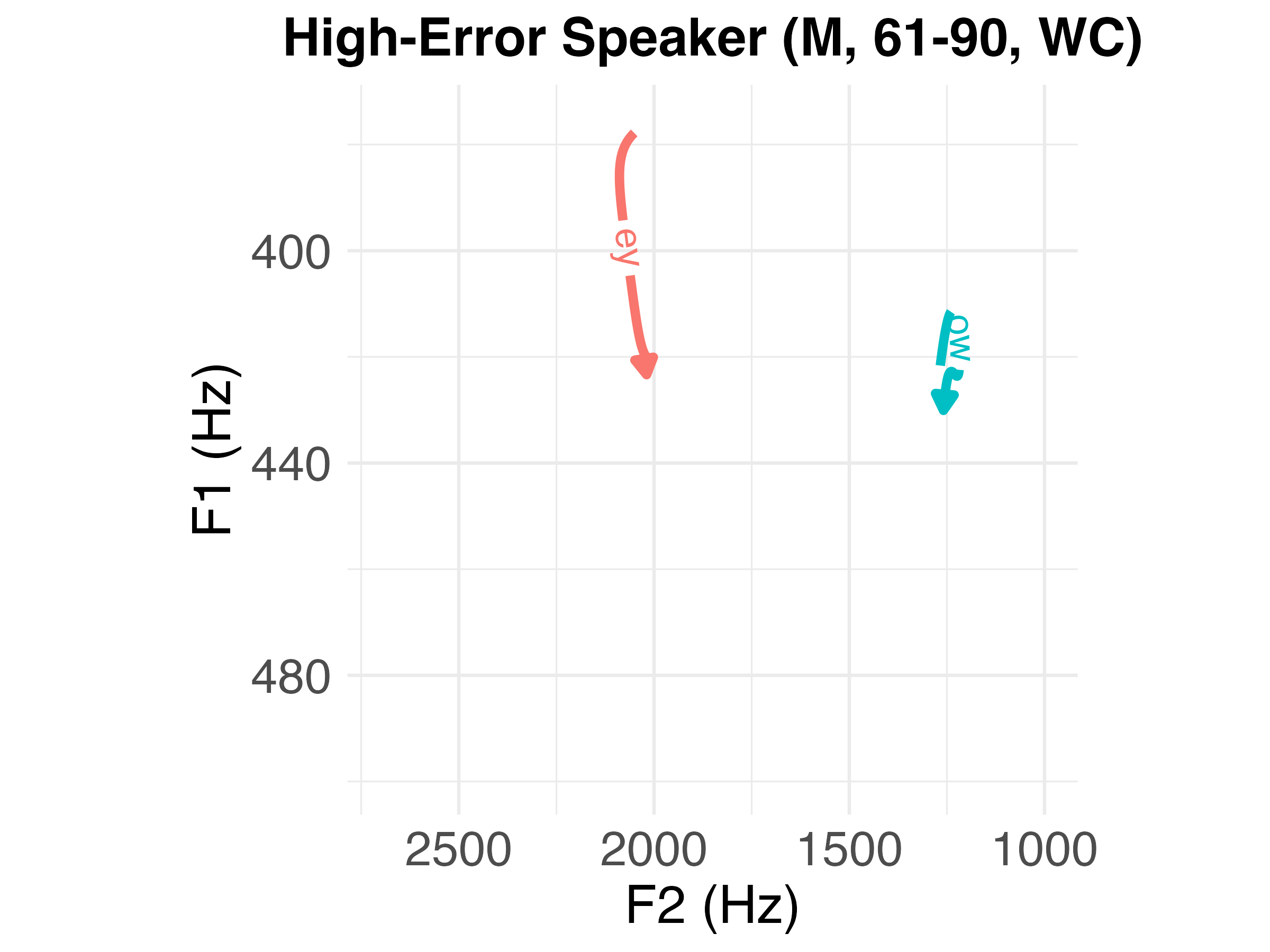}
\end{subfigure}
\caption{Formant trajectories of speakers with the highest number of errors.}
\label{fig:VirginiaLawrence}
\end{figure}

Figure \ref{fig:VirginiaLawrence} shows two speakers of the same age and socioeconomic background who also received a notable amount of errors $(n_{\text{man}} = 6;\ n_{\text{woman}} = 14)$. Only 4 of these had to do with the \textsc{goat} vowel, 2 errors per person. In this instance, we can see that the woman, who received the most errors, has a shortened glide, but not a centering diphthong. The man's glide is longer, but the direction points to the center. As for the \textsc{goat} vowel, the woman once again has a more mainstream realisation than the man, even though both glides are short compared to, e.g., the speaker in Figure \ref{fig:decten2y10i026}.

\begin{figure}[H]
\centering
\begin{subfigure}{0.52\textwidth}
    \centering
    \includegraphics[width=\linewidth]{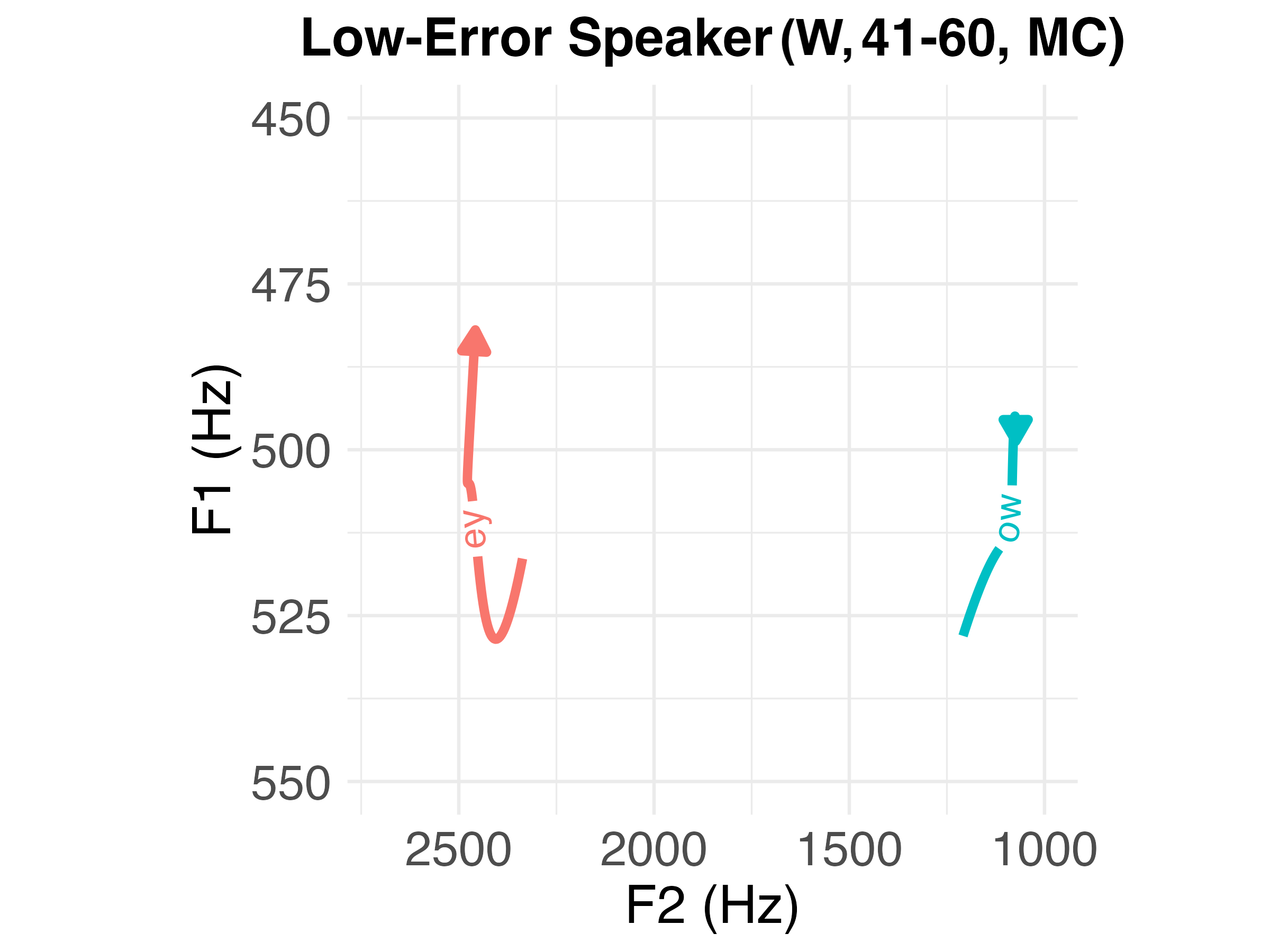}
\end{subfigure}%
\begin{subfigure}{0.52\textwidth}
    \centering
    \includegraphics[width=\linewidth]{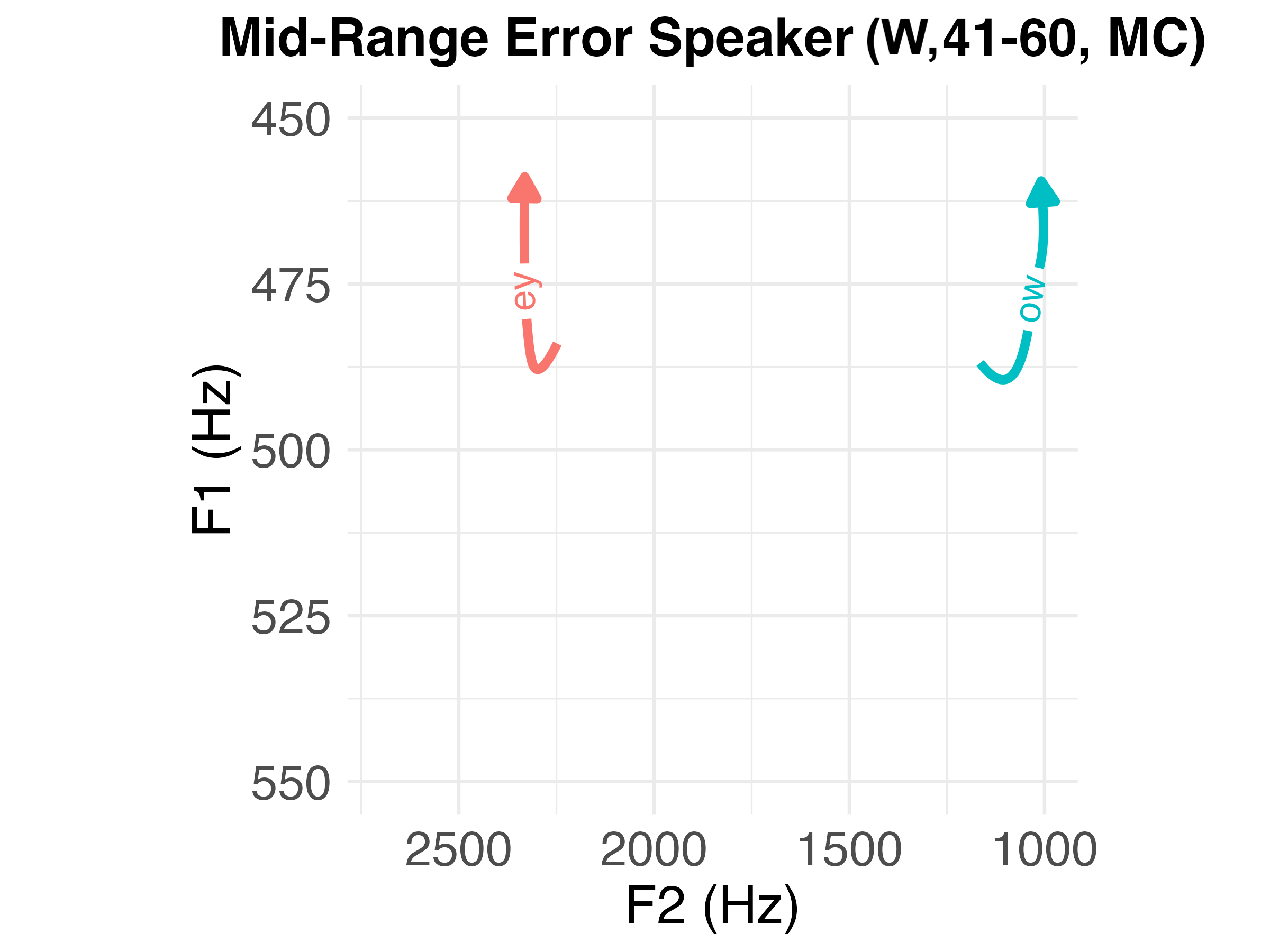}
\end{subfigure}
\caption{Formant trajectories of speakers in the Mid-range error group.}
\label{fig:BarbaraCynthia}
\end{figure}

Figure \ref{fig:BarbaraCynthia} represents the formant trajectories of two women of the same age and socioeconomic background. Interestingly, one of them (Figure \ref{fig:BarbaraCynthia}, right) received 5 out of the total 6 errors, even though their sociolinguistic background is quite similar. Looking at their realisations of the \textsc{face} and \textsc{goat} vowels, they certainly differ in vowel height, but the glide trajectory is quite similar. The speaker on the left, who received just one error, has more pronounced glides. Overall, their formant trajectories look rather mainstream compared to the speakers who received more errors.

The acoustic analyses provide some descriptive evidence that speakers with more \textsc{face}-related ASR errors tend to show reduced formant movement, consistent with more monophthongal realisations. No comparable pattern was observed for \textsc{goat}. The individual case studies further illustrate variation in glide reduction and centering quality across speakers. Given the small acoustic sample, future research would benefit from combining sociolinguistic and acoustic approaches across a larger number of speakers to investigate the phonetic factors contributing to ASR errors.

\section{Discussion}
\label{sec5}

In this paper, we confirm our earlier findings \citep{serditova25_interspeech} regarding gender, with men receiving more errors than women. The findings of previous research are mixed regarding the direction of the bias \citep[see][for a review]{ngueajio2022hey}. \citet{liu2022towards} report higher word error rates for men across several ASR systems, even after fine-tuning with in-domain data. \citet{garnerin2021investigating} evaluated the impact of the ratio of men and women in the training data on the ASR performance and found that women consistently had higher WER than men, even though the model was trained on substantially more speech from women than from men. Gender-related performance gaps may arise from multiple interacting factors, including the composition and representativeness of training data as well as architectural and representational properties of ASR models, which may help explain why such disparities are not necessarily eliminated through additional fine-tuning. Using sociolinguistic frameworks for the analysis, we argue that the poorer ASR performance on speech from men in our study is linked to men using more non-standard forms. Indeed, sociolinguistic research has consistently shown that women tend to favour more standard forms \citep{trudgill1972sex,labov1963social,nichols1983linguistic}. Non-standard forms are also processed less quickly and less accurately than standard forms by listeners who use standard varieties in a wide range of tasks from vowel identification to semantic processing \citep{ClopperPerceptionDialectVariation_2021}. Similar to listeners who speak a standard variety, ASR is trained on mainstream data containing speech of standard varieties, it is not surprising that non-standard forms are harder to process and more vernacular speakers experience poorer performance. 

Among the other social factors considered, age did not significantly predict ASR error counts, nor did it significantly interact with error type, indicating no reliable age-related differences in error patterns in the present data. Descriptively, however, the ASR system produced more errors for younger and older speakers than for working-age adults. This numerical pattern is compatible with age grading, a sociolinguistic pattern in which speech changes across the lifespan. Younger speakers often use more non-standard forms as linguistic innovators, whereas adults in the workforce tend to shift toward standard variants under professional, social, and educational pressures. Later in life, as these pressures diminish, speakers often reintroduce vernacular features acquired in adolescence. Age grading has been documented in Newcastle \citep{grama2023tracking, grama2023post, bauernfeind2023change, moelders2025navigating}. However, because the age effect did not reach statistical significance in the present analysis, we cannot conclude that age grading accounts for the observed descriptive pattern; this remains a hypothesis for investigation with a larger sample.

Both of these findings indicate that ASR biases echo the sociolinguistic patterns that exist in human-human communication, as well as reinforce real-world biases and stigmas. This is especially evident when we look at dialectal errors and the linked local linguistic features on a case-by-case basis and analyse them based on the local socioeconomic climate and history. This yields more nuanced insights into how technological biases intersect with regional identity and social inequality. We know the value of this approach from the second wave of sociolinguistics. It focused on ethnographic methods and more nuanced and local contexts \citep{milroy1980language}, a step further than treating speakers as ``bundles of demographic characteristics'' during the first wave of sociolinguistics \citep{labov1972language}. Assessment of ASR bias seems to currently follow the first-wave sociolinguistic tradition, often relying on overly broad demographic categories such as ethnoracial affiliation, gender, or age \citep{BeraAgarwal2025,ngueajio2022hey,Nguyen-CollaborativeGrowth-2025}. While these categories are undoubtedly practical, they might obscure the complexity of how speakers actually use language in real-world contexts \citep{markl-etal-2024-language}. In this research, we not only focused on a specific regional community -- Newcastle -- but also dissected ASR errors one by one to demonstrate that they are directly linked to local linguistic features. While it does presuppose a more narrow approach, the strength is that we were able to assess ASR biases on a community level, applying the same lens that sociolinguists use to study language variation and identity. We were able to show that ASR systems are not just biased in abstract, statistical terms, but in very concrete ways that reflect their failure to accommodate linguistic diversity.

We demonstrated that phonological and lexical errors are the two leading types of ASR errors. This is not surprising: certain features, particularly phonological ones, are more frequent and indexical of local identity, and their occurrence is more likely in spontaneous, naturalistic speech. Compared to them, morphosyntactic features are less frequent. Interestingly, this trend is also reflective of real-world sociolinguistics. Syntactic variation is notoriously under-explored compared to sociophonetic features, which is due to lower frequency of syntactic constructions, as well as the challenge of accounting for the semantic and pragmatic meanings that syntactic constructions usually possess \citep{moore2021social,moore2023socio, serditova2025meet}. In this study, we demonstrated that WER for Newcastle speakers is high due to several factors. First, the abundance of phonological errors means that the acoustic model is not trained on enough data that reflects regional phonological variation. Second, the large numbers of lexical errors might signify that these regional lexical items are simply absent from the model's dictionary. Unfortunately, we can only speculate about this, since Rev AI's full dictionary is not publicly accessible.

As for morphosyntactic errors, one should keep in mind their connection to standardisation errors. In fact, the majority of standardisation errors in our dataset involve morphosyntactic features. Out of 294 errors in this error type, 38 are related to vocabulary, but the rest are linked to pronouns, verb paradigm, and the local usage of plural forms. Thus, it is the local syntactic constructions or forms that are most likely to undergo standardisation by ASR. This tendency likely stems from the fact that morphosyntactic forms often deviate from standard patterns in subtle but regular ways that ASR systems are not well-trained to recognise \citep{martin2020understanding}. Unlike distinctive lexical items, which may be learned as individual exceptions, or phonological variants, which may be acoustically ambiguous, morphosyntactic variation tends to be systematic yet underrepresented in ASR training data \citep{koenecke2020racial,MartinCorpusData2021}. This makes it especially vulnerable to standardisation. The system ``corrects'' what it perceives as ungrammatical or anomalous, which reinforces mainstream language norms.

Our analysis also revealed previously undocumented patterns. In the final model, we observed a significant interaction between error type and gender. Post-hoc analyses using estimated marginal means revealed that the significant interaction between error type and gender was primarily driven by differences in lexical errors. Men showed substantially higher predicted counts of lexical errors than women ($\hat{\mu} = 9.12$, 95\% CI [6.94, 11.99] vs.\ $\hat{\mu} = 5.25$, 95\% CI [3.86, 7.14]). In contrast, phonological errors were the most frequent error type for both genders, but showed only a modest gender difference ($\hat{\mu} = 24.58$ for men, $\hat{\mu} = 23.05$ for women). Standardisation and morphosyntactic errors exhibited minimal gender differences, with overlapping confidence intervals (standardisation: $\hat{\mu} = 4.10$ for men, $\hat{\mu} = 4.19$ for women; morphosyntactic: $\hat{\mu} = 4.05$ for men, $\hat{\mu} = 3.28$ for women). This, too, points out to men being more likely to use regionally distinctive forms (particularly lexical items), which the ASR system struggled to process, thereby contributing disproportionately to the error rates in these categories.

Socioeconomic status did not significantly improve model fit and was therefore excluded from the final model, indicating that SES was not a reliable predictor of ASR error counts in the present dataset. This may reflect low statistical power due to unbalanced group sizes. Overall, while class remains a salient sociolinguistic factor in the UK, such distinctions may be less pronounced in a predominantly working-class city like Newcastle. Nevertheless, the descriptive differences observed across SES groups suggest that the relationship between socioeconomic variation and ASR performance warrants further investigation in a larger and more socioeconomically balanced sample.

An important limitation of the present study concerns the interpretation of ASR performance disparities. Previous work has shown that the intelligibility of Geordie speech depends on listener familiarity and linguistic background rather than solely on properties of the speech signal itself \citep{du2023rapid}. Consequently, the performance disparities observed in the present study should not be interpreted as evidence of a single source of ASR bias, as our analyses do not compare ASR performance with human intelligibility for the same speech material. Future work combining ASR evaluation with human perception experiments would help disentangle these effects.

\subsection{Dialectal Bias}
\label{subsec5.1}

While in many cases the root of error, i.e. the regional linguistic feature that is clearly responsible for the ASR mishap, is obvious, there are instances where several features contribute to one error. For example, in the error ``wor lass gans'' → ``we were asking'', we can see three regional features: (1) the local pronoun ``wor'', which we showed to be a struggle for ASR; (2) the local lexical item ``lass''; and (3) part of the local verb paradigm ``gan'' (meaning ``go''). Or, this is an example with two phonological features: ``quaint'' → ``queen''. In this case, we can see that both the \textsc{face}-monophthongisation and the glottal stop impacted the ASR decision. In our analysis, we counted these as separate errors because we wanted to look at the root cause. However, we can see many arguments for the opposite approach. Grouping such compound errors into single units might better reflect the perceptual reality of ASR processing, where multiple interacting features jointly lead to misrecognition. It also raises the question of error granularity — whether we should prioritise the linguistic root cause of error (e.g., phonology vs. lexicon) or the surface output, which may combine several regional cues. Ultimately, our choice to separate them analytically allowed for a clearer identification of which features most consistently challenge ASR, but future work might benefit from a more integrated categorisation that captures the interaction of features.

This observation aligns with findings by \citet{markl2022language}, who notes that substitution errors can arise from either phonetic similarity (with no semantic relation) or morphological relatedness (without phonetic resemblance). In our data, we encountered errors that involved both these types, e.g., ``quaint'' → ``queen'', ``wor'' → ``a'' (in the context ``And then keep the rest of it for \textit{wor} (ground truth) /\textit{a} (ASR output) bus fare''). These compound substitutions show the complexity of regional dialects and the challenge of disentangling phonetically driven misrecognitions from those influenced by morphosyntactic, semantic or lexical expectations.

We also demonstrated the acoustic patterns behind ASR misrecognitions, based on the vowels \textsc{face} and \textsc{goat}. There is evidence that ASR struggles more to provide a correct output for the speakers whose realisation of these vowels is regional, be it a shortened glide or a centering quality to these diphthongs. Thus, acoustic variation remains a challenge for ASR systems. It is yet another proof that ASR models are not sufficiently trained on the full spectrum of sociophonetic variation found in natural speech, particularly that which departs from the prestige norm.

\subsection{Sociolinguistics of Speech Technology}
\label{subsec5.2}

\citet{kelly2025artificial} argues that AI systems, including speech technologies, introduce significant sociolinguistic biases, particularly in areas such as automatic dialect and accent identification. The author discusses the concept of algorithmic identity, which is the idea that AI systems tailor linguistic behaviour based on users' demographic and behavioural data. The notion helps explain why ASR performance differs across dialect groups, as ASR systems are optimised for the ``algorithmic identities'' they most frequently encounter (typically white, middle-class, mainstream language users). On the other hand, marginalised dialect users are algorithmically under-represented, receiving, for instance, poorer ASR performance. Thus, we can frame dialectal bias not just as a technical failure, but as a sociolinguistic hierarchy built into training data.

In response, \citet{kessler2024socio} emphasise that sociolinguists should actively shape, rather than merely react to, the development of generative AI technologies. While technological change is inevitable, it also creates new research opportunities, such as large-scale corpus analysis, netnography, and AI-assisted annotation, provided that human validation and ethical reflection remain central. One concrete avenue for such intervention lies in the systematic annotation of dialectal variation. Sociolinguists can contribute by developing dialectal feature taggers that capture both phonological variation \citep[e.g.][]{MojaradTang_2025_ASRAAE_Interspeech,Kendalletal_ING_2021} and morphosyntactic variation \citep[e.g.][]{Harrison_Martin_Moeller_Tang_Disambig_Hab_be_2022,PrevilonRozetGowdaDyerTangMoeller_LREC_Accepted_2024,Johnson2024,porwal2025analysisllmgrammaticalfeature, Nguyen-CollaborativeGrowth-2025}, and by incorporating these annotations into ASR pipelines. These features can be used not only to improve training data, but also, as demonstrated in this paper and a handful of other studies such as \citet{wassink2022uneven}, to provide more fine-grained evaluation of ASR performance beyond broad categories such as race, gender and age.


The importance of sociolinguistic input in improving speech technologies is also highlighted by \citet{mallinson2024place}, who show how sociolinguistic expertise can contribute to areas such as deepfake detection, listener perception, and the creation of more representative speech datasets. Likewise, \citet{grieve2025sociolinguistic} argue that language technologies can only function equitably when their training data reflect the full sociolinguistic diversity of the language varieties they model.

As \citet{dong2024fairness} argues, AI systems reproduce social biases and are increasingly embedded in social interaction. Sociolinguists should examine not only technological performance, but also how AI-mediated communication modifies the notions of social meaning, identity, and interaction, and introduces new forms of subjectivity in research. Beyond improving technology, sociolinguistic methods can be used to study speech technologies themselves. Research shows that AI systems do not only process language but also influence language variation and change. For example, \citet{szekely2025will} demonstrate that synthetic voices carry socioindexical cues and that users accommodate to conversational agents, while ASR systems reproduce biases present in their training data. \citet{foster2023social} argue that human–robot interaction is constrained by the same social and linguistic factors as human–human communication, including accent, dialect, style-shifting, and identity cues. These developments show that speech technologies should be treated as sociolinguistic actors rather than neutral tools, whose performance and biases provide insight not only into technology itself, but also into more general patterns of human communication and social structure.

Our findings also point to specific priorities for improving dialect-aware ASR. Rather than focusing only on overall WER disparities across broad demographic groups, developers could use fine-grained linguistic error analyses to identify the particular features that systematically trigger misrecognition. In the present study, phonological variation was the primary source of dialect-related errors, particularly vowel quality, glottalisation, and monophthongisation, while local vocabulary and morphosyntactic forms such as regional pronouns and verb paradigms also presented recurrent difficulties. These findings suggest that training and evaluation data should not only include more speakers of underrepresented dialects, but should adequately represent the specific phonological, lexical, and morphosyntactic variation found within these communities.

One of the main goals of this paper was to demonstrate that ASR biases reflect real-world social inequalities, not only across broad categories such as gender or age, but also across regional and dialectal variation. Speakers of non-standard varieties may therefore be systematically misunderstood by technologies increasingly used in education, employment, healthcare, and public services (see Section~\ref{sec1}), effectively marginalising linguistic identities that fall outside the standard language norm. This pattern is closely tied to the limited representation of regional speech in ASR training data, which often privileges mainstream varieties. We argue that sociolinguistic awareness must be incorporated into the development and evaluation of speech technologies if they are to perform equitably across diverse speech communities \citep{Choi_Choi_2025}, echoing calls to centre the needs and perspectives of language communities in technological development \citep{markl-etal-2024-language}.

\section{Conclusion}

This study examined patterns of Automatic Speech Recognition (ASR) errors in sociolinguistic interview data, focusing on phonological, lexical, morphosyntactic, and standardisation errors across speaker gender, age, and socioeconomic background. A negative binomial mixed-effects model revealed significant differences by error type and gender, with lexical errors particularly elevated among men, demonstrating that ASR performance can reflect underlying sociolinguistic variation. These findings show that ASR errors are not distributed randomly but are systematically associated with socially and regionally patterned ways of speaking, rather than demographic categories alone. We therefore argue that ASR evaluation must move beyond aggregate accuracy measures and incorporate sociolinguistic variation, as recognising regional and social diversity is essential for developing fairer, more inclusive speech technologies.

\section{Acknowledgements}
We acknowledge the Hochschulinterne Forschungsförderung (HiFF) at Hochschule Düsseldorf. Together with Prof. Jochen Steffens and our research assistants Paula Thees and Vincent Reichmann, this support made the work possible as a collaborative team effort. We especially thank Paula Thees and Vincent Reichmann for their careful data processing and technical support. We are also grateful to Prof. Karen Corrigan for granting us access to the DECTE corpus and to Prof. Isabelle Buchstaller and Dr. James Grama for providing us access to the \emph{Language Change Across the Lifespan} project files.

\section{Author contributions: CRediT}
We follow the CRediT taxonomy\footnote{\url{https://credit.niso.org/}}. Conceptualisation: DS, KT; Data curation: DS; Formal Analysis: DS; Funding acquisition: KT, DS; Investigation: DS; Methodology: DS, KT; Resources: KT; Software: KT, DS; Visualisation: DS, KT; and Writing – original draft: DS, review \& editing: DS, KT.

\section{Data availability}
The data and scripts necessary to reproduce the results presented are available in an Open Science Framework repository, \url{https://osf.io/42n9r/}.

\section{Declaration of generative AI and AI-assisted technologies in the manuscript preparation process.}
During the preparation of this work the author(s) used ChatGPT for language refinement and readability enhancement. After using this tool/service, the author(s) reviewed and edited the content as needed and take(s) full responsibility for the content of the published article.

\appendix

\section{Acoustic Analysis: Remaining Speakers}
\label{appendix}

Formant trajectories of the remaining speakers analysed in Section~\ref{subsec:acousticanalyses}. Please mind the y-axis differences.

\begin{figure}[H]
\begin{flushleft}
    \centering
    \includegraphics[width=0.6\textwidth]{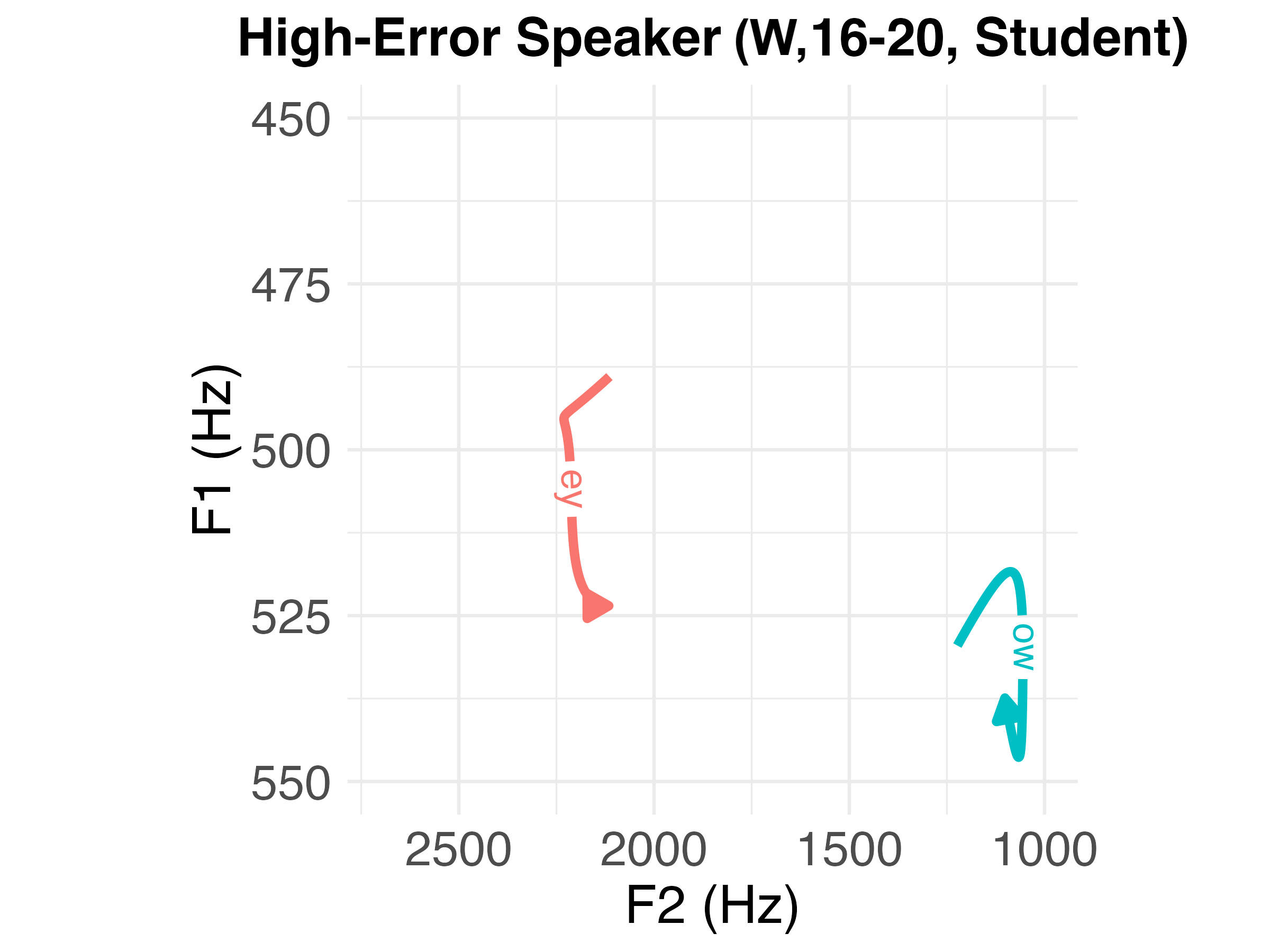}
    \caption{Formant trajectory of a speaker with a high number of errors.}
    \label{fig:decten2y10i018b}
\end{flushleft}
\end{figure}

\begin{figure}[H]
\centering
\begin{subfigure}{0.52\textwidth}
    \centering
    \includegraphics[width=\linewidth]{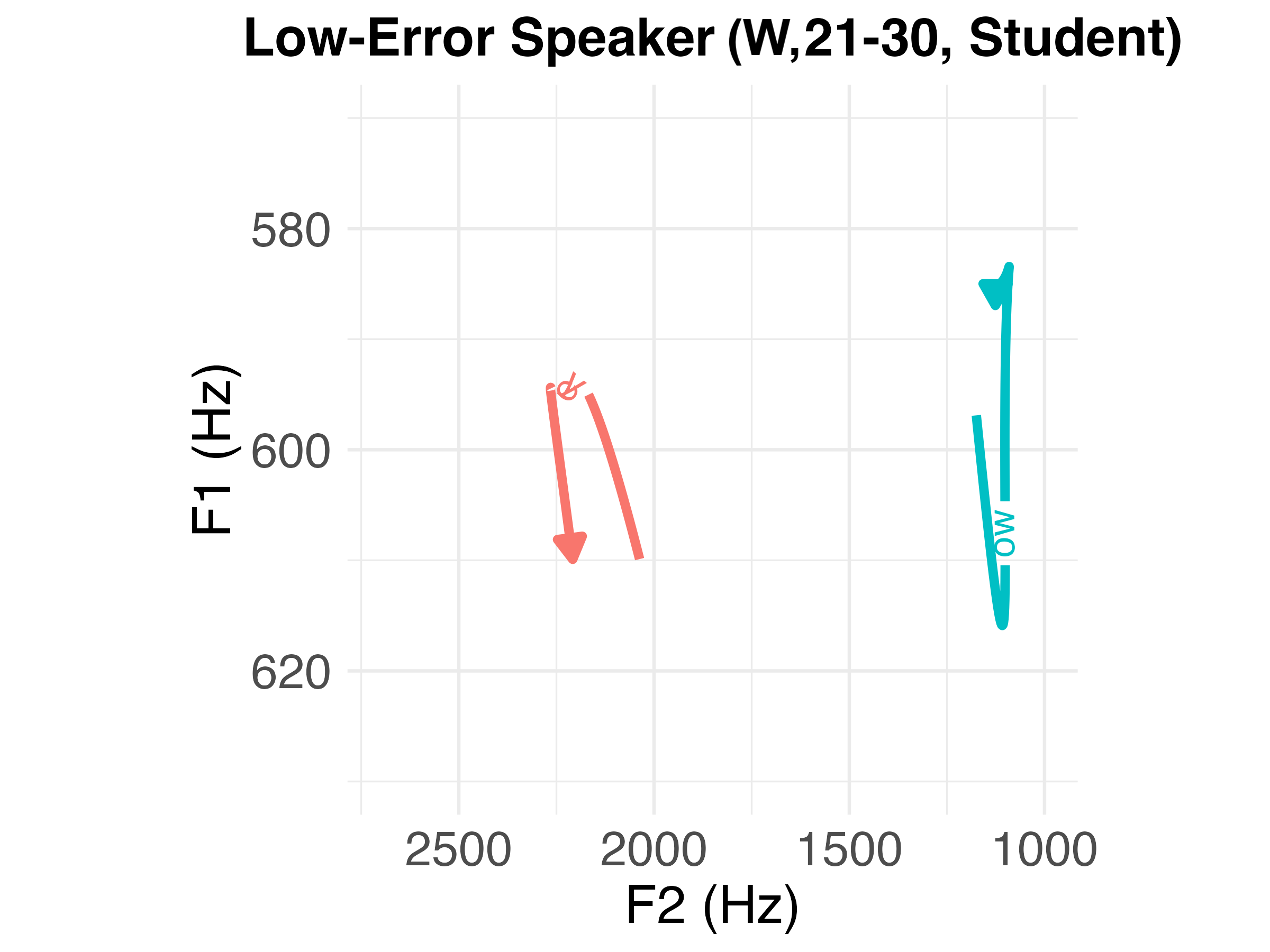}
\end{subfigure}%
\begin{subfigure}{0.52\textwidth}
    \centering
    \includegraphics[width=\linewidth]{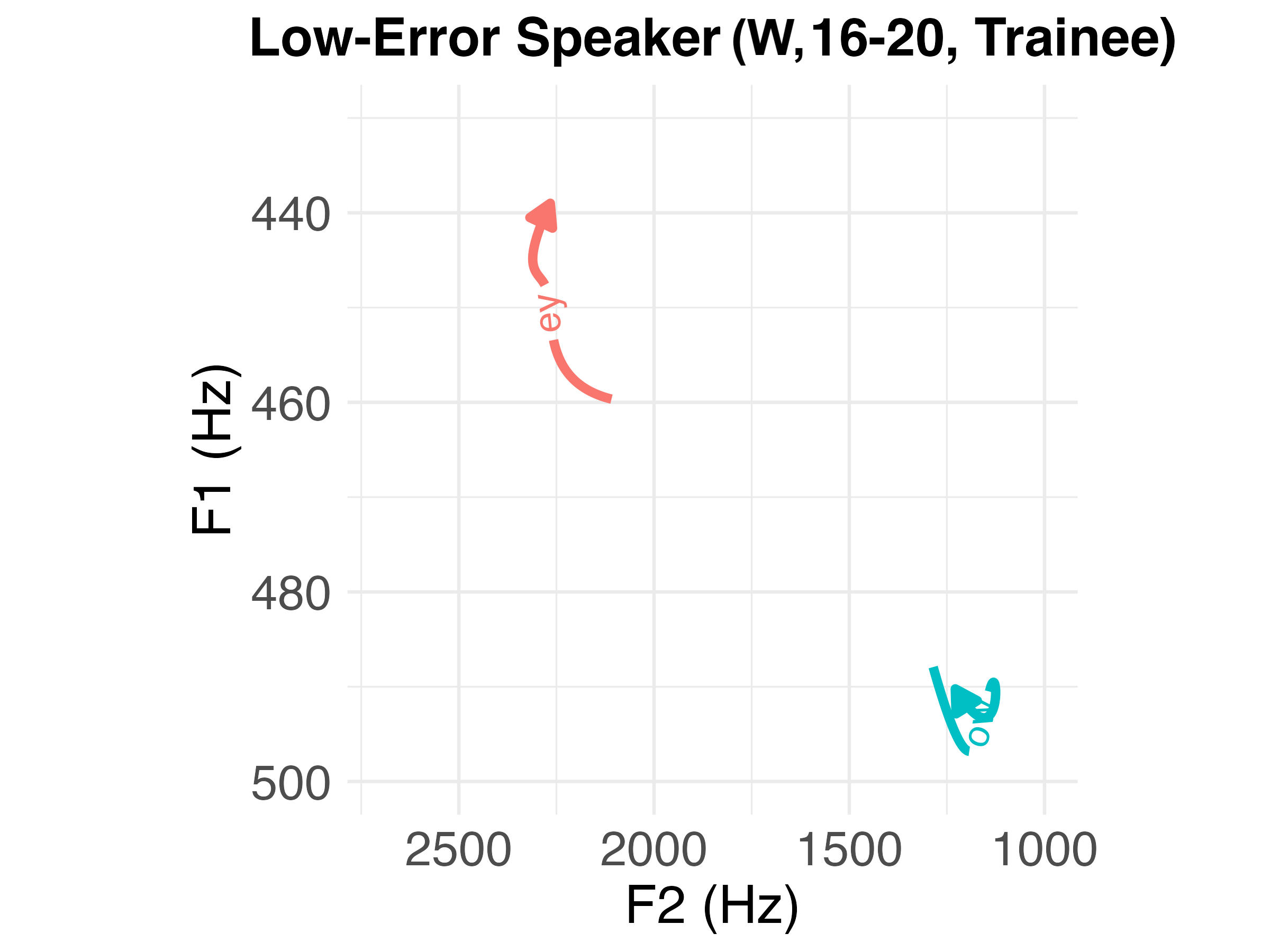}
\end{subfigure}
\caption{Formant trajectories of speakers with a low number of errors.}
\label{fig:decten2y10i018aAngela}
\end{figure}

\begin{figure}[H]
\centering
\begin{subfigure}{0.52\textwidth}
    \centering
    \includegraphics[width=\linewidth]{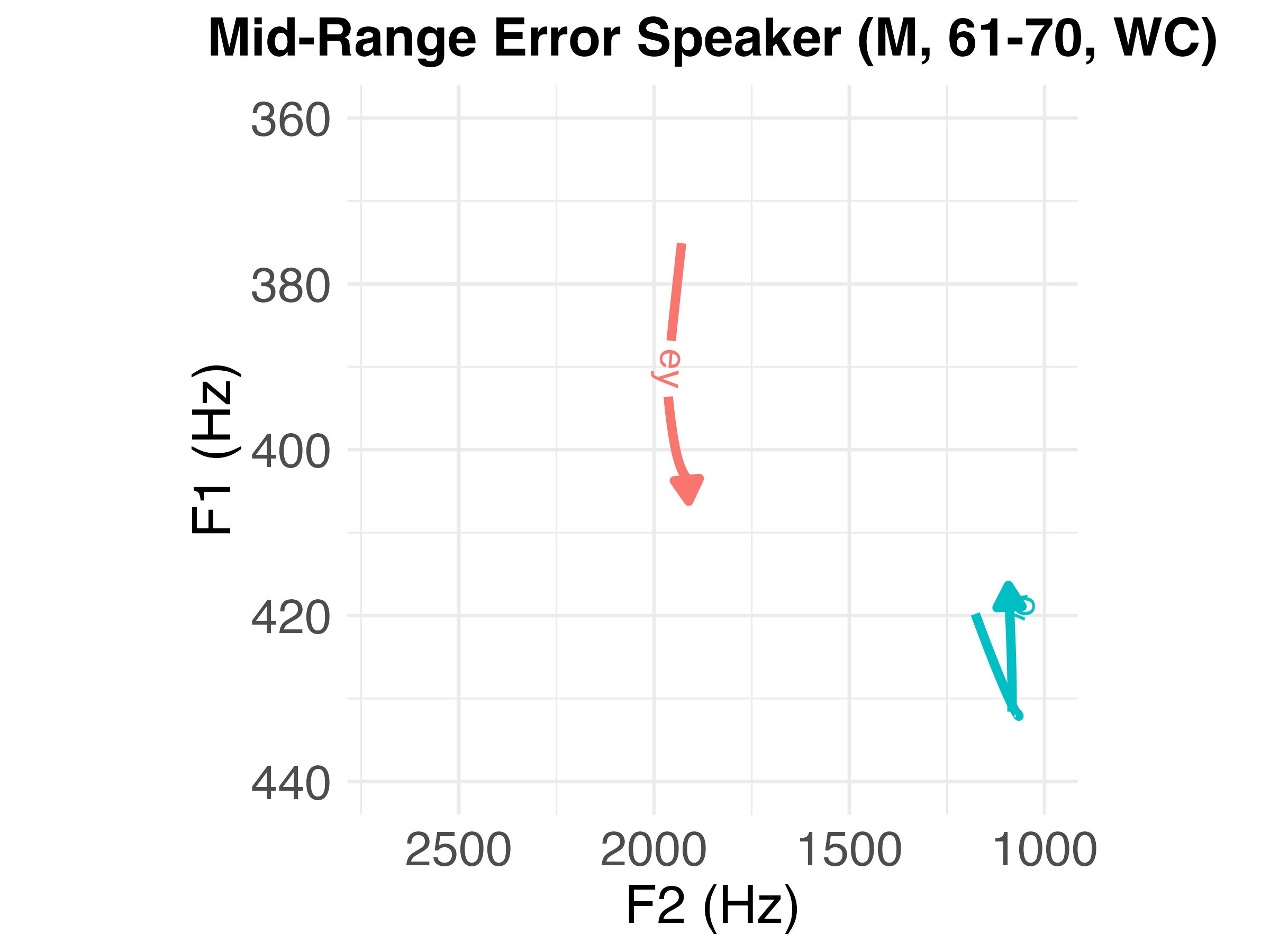}
\end{subfigure}%
\begin{subfigure}{0.52\textwidth}
    \centering
    \includegraphics[width=\linewidth]{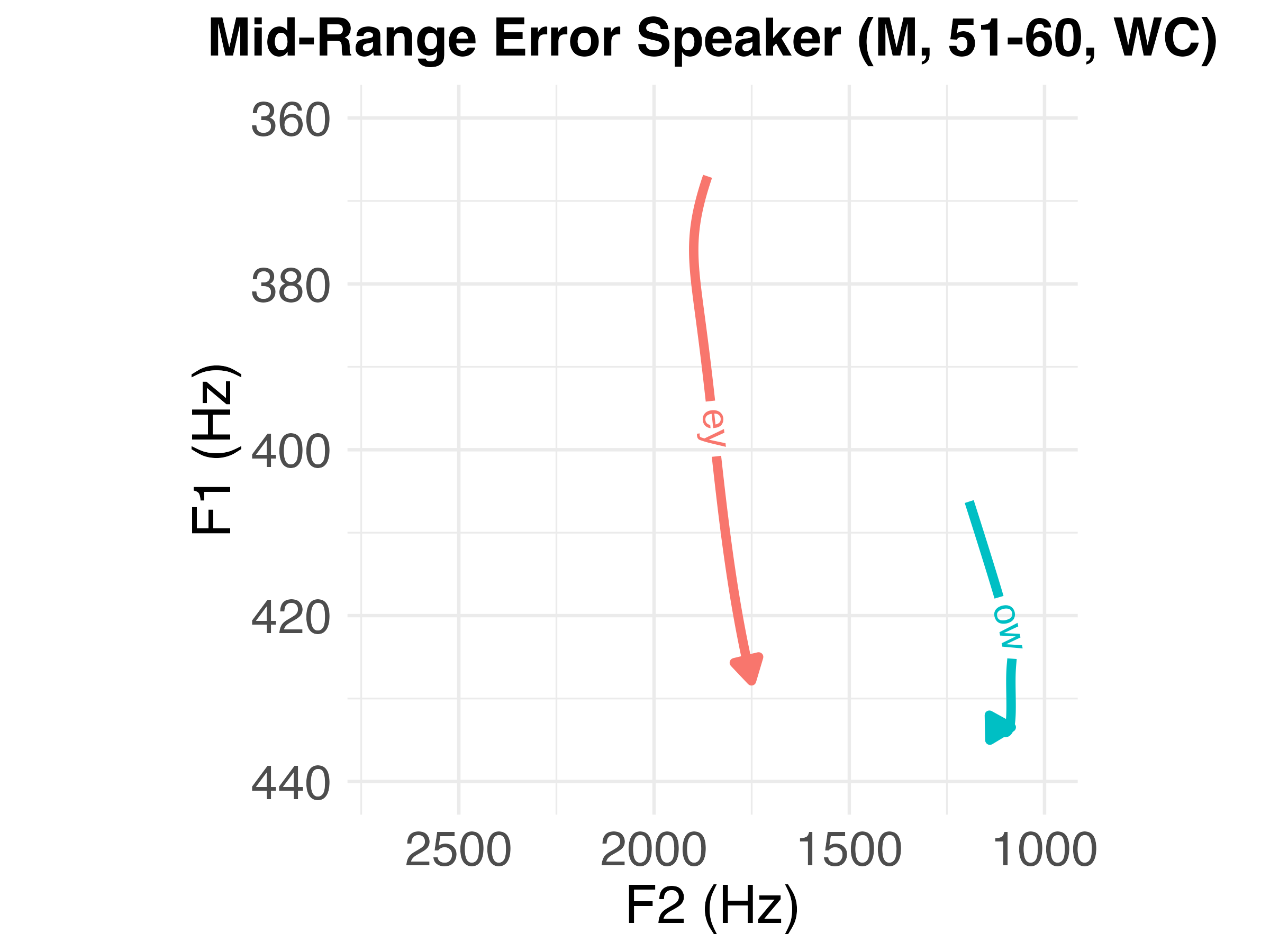}
\end{subfigure}
\caption{Formant trajectories of speakers in the Mid-range error group.}
\label{fig:HenryRaymond}
\end{figure}

\begin{table}[ht]
\centering
\caption{Distribution of speakers included in the analysis by age, gender, and socioeconomic status.}
\label{tab:participants}
\begin{tabular}{llrr}
\toprule
\textbf{Variable} & \textbf{Category} & \textbf{N} & \textbf{\%} \\
\midrule
\multirow{4}{*}{Age}
& 16--20 & 19 & 23.5 \\
& 21--40 & 18 & 22.2 \\
& 41--60 & 22 & 27.2 \\
& 61--90 & 22 & 27.2 \\
\midrule
\multirow{2}{*}{Gender}
& Men & 42 & 51.9 \\
& Women & 39 & 48.1 \\
\midrule
\multirow{7}{*}{Socioeconomic status}
& Homemaker & 4 & 4.9 \\
& Working class & 14 & 17.3 \\
& Middle class & 19 & 23.5 \\
& Retired & 17 & 21.0 \\
& Student & 23 & 28.4 \\
& Unemployed & 3 & 3.7 \\
& Unknown & 1 & 1.2 \\
\bottomrule
\end{tabular}
\end{table}

\bibliographystyle{apalike} 
\bibliography{Tang}

\end{document}